\documentclass{article} % For LaTeX2e
\usepackage{arxiv}
\usepackage[utf8]{inputenc}
\usepackage[T1]{fontenc}
\usepackage{microtype}
\usepackage{xcolor}
\usepackage[numbers,square,sort&compress]{natbib}

\usepackage{amsmath,amsfonts,bm}

\def\eqref#1{equation~\ref{#1}}
\def\1{\bm{1}}

\DeclareMathAlphabet{\mathsfit}{\encodingdefault}{\sfdefault}{m}{sl}
\SetMathAlphabet{\mathsfit}{bold}{\encodingdefault}{\sfdefault}{bx}{n}

\usepackage{hyperref}
\usepackage{url}
\usepackage{graphicx}
\usepackage{algorithm}
\usepackage{algorithmic}
\usepackage{mathtools}

\usepackage{booktabs}
\usepackage{array}
\usepackage{amsmath}
\usepackage{amssymb}
\usepackage{wrapfig}
\usepackage{subcaption}

\title{Prioritized Rollouts for Efficient 
World Model-based Vision-Language-Action Policy Optimization}
\renewcommand{\shorttitle}{Prioritized Rollouts for Efficient 
World Model-based Vision-Language-Action Policy Optimization}
\date{}
\author{
\begin{tabular}{@{}l@{}}
\textbf{
Yifei Sheng\textsuperscript{1,2,3,*}
\quad
Haoxiang Ren\textsuperscript{1,2,3,*}
\quad
Zhilong Zhang\textsuperscript{1,2,*}
\quad
Haonan Wang\textsuperscript{1,2,3,*}
}
\\
\textbf{
Runjie Xu\textsuperscript{1,2,3}
\quad
Yihao Sun\textsuperscript{4,5}
\quad
Nan Tang\textsuperscript{1,2,3}
\quad
Zhichao Wu\textsuperscript{1,2}
\quad
Lei Yuan\textsuperscript{1,2}
}
\\
\textbf{
Haoxin Lin\textsuperscript{1,2,3}
\quad
Yang Yu\textsuperscript{1,2,\textdagger}
}
\end{tabular}
\\[0.8em]
\textnormal{
\textsuperscript{1}National Key Laboratory for Novel Software Technology,
Nanjing University, Nanjing, China
}
\\
\textnormal{
\textsuperscript{2}School of Artificial Intelligence,
Nanjing University, Nanjing, China
}
\\
\textnormal{
\textsuperscript{3}Cirquar Technologies, Nanjing, China
\qquad
\textsuperscript{4}Mila -- Quebec AI Institute
}
\\
\textnormal{
\textsuperscript{5}Universit\'e de Montr\'eal
}
}

\newcommand{\method}{\textsc{U-GROW}}
\newcommand{\Doff}{\mathcal{D}_{\mathrm{off}}}
\newcommand{\Dwm}{\mathcal{D}_{\mathrm{wm}}}
\newcolumntype{C}[1]{>{\centering\arraybackslash}p{#1}}

\begin{document}

\maketitle

\begingroup
\renewcommand{\thefootnote}{\fnsymbol{footnote}}
\footnotetext[1]{Equal contribution.}
\footnotetext[2]{Corresponding Author: \href{mailto:yuy@nju.edu.cn}{yuy@nju.edu.cn}}
\endgroup

\begin{abstract}
Vision-Language-Action (VLA) models have emerged as a powerful paradigm for embodied intelligence, but fine-tuning them with reinforcement learning (RL) remains constrained by the cost of real-world robot interaction. Model-based reinforcement learning (MBRL) reduces this cost by using a learned world model to generate rollouts for policy optimization. However, it becomes computationally expensive as VLA policies and world models scale. Existing methods typically treat states equally, overlooking substantial differences in their utility for policy improvement. In this paper, we show that policy uncertainty helps identify states with greater potential for policy improvement. 
The policy exhibits high uncertainty at only a small subset of states, often during decision-sensitive stages where small action differences can alter task outcomes, 
suggesting that policy improvements at these states could be particularly valuable.
Building on these findings, we introduce \textbf{\textsc{U-GROW}}, a lightweight, plug-and-play sampling layer that directs more model rollouts to these informative states. By modifying only the branched-start distribution, \textsc{U-GROW} can be integrated into existing MBRL pipelines without changing the policy optimization objective. Experiments in both simulated and real-world manipulation tasks demonstrate the efficiency and effectiveness of \textsc{U-GROW}, supporting the use of policy uncertainty to guide experience generation.
\end{abstract}

\section{Introduction}
Recent advances in Vision-Language-Action (VLA) models have marked substantial progress toward general-purpose robotic control~\citep{kim2024openvla,intelligence2025pi}. Building on the broad capabilities learned through offline imitation, reinforcement learning (RL) can further improve the task performance and generalization of VLA policies~\citep{chen2025pirl, li2026simplevla}. However, online RL on physical robots requires repeated real-world robot interaction, which is slow, costly, and difficult to parallelize. World models can alleviate this bottleneck by generating rollouts for policy optimization, thereby reducing the need for additional real-robot interaction~\citep{yu2026should,hou2026world}. Recent methods following this paradigm have demonstrated effective RL fine-tuning of VLA policies using imagined rollouts~\citep{zhu2026wmpo,zhang2026vlambpo}.

As VLA policies and world models scale~\citep{zhang2024whale,agarwal2026cosmos}, the computational costs of rollout generation and policy optimization increase~\citep{zhu2026wmpo,zhang2026vlambpo,guo2026vlaw}, making efficient use of model-generated experience essential.
Intuitively, states may differ in their value for policy improvement. For example, correcting a small alignment error may turn failure into success, whereas further optimizing already reliable free-space motion may yield little additional benefit.
However, existing methods typically do not explicitly account for these state-level differences. This raises a central question: 
% \begin{center}
\emph{Which states offer the greatest potential for policy improvement, and how can they be identified?}
% \end{center}

Policy uncertainty may help identify such states.
High uncertainty can indicate less reliable action predictions~\citep{ren2026when}, suggesting potential room for policy improvement. 
% Our empirical analysis shows that high-uncertainty states are sparse and tend to cluster at decision-sensitive manipulation stages, such as grasping, placement and release, where small differences in spatial alignment or timing can alter subsequent interactions and task outcomes. 
Our empirical analysis shows that the policy exhibits high uncertainty at only a small subset of visited states, often during decision-sensitive manipulation stages such as grasping, placement, and release.
At these stages, small differences in spatial alignment or timing can alter subsequent interactions and task outcomes.
More directly, a policy-switching intervention shows that, at comparable replacement rates, replacing the base policy with a stronger policy at high-uncertainty states yields substantially higher success rates than doing so at low-uncertainty states.
Together, these findings support using policy uncertainty to identify states with greater potential for policy improvement.

Motivated by these observations, we propose \method{} (\textbf{U}ncertainty-\textbf{G}uided \textbf{R}ollouts for policy \textbf{O}ptimization with \textbf{W}orld models), a lightweight method that directs model rollouts toward states with greater potential for policy improvement. 
Specifically, we adopt short-horizon branched rollouts~\citep{janner2019mbpo} and use an uncertainty-guided sampling layer to select their starting states.
After filtering out difficult-to-recover failure tails, this layer prioritizes high-uncertainty states as rollout starts. 
By modifying only the rollout-start distribution, \method{} can be plugged into existing MBRL pipelines without changing their policy optimization objectives. 
We evaluate \method{} on the RoboTwin 2.0 and LIBERO benchmarks, as well as four real-world manipulation tasks. 
Across these settings, \method{} improves both sample efficiency and final success rates over uniform rollout-start sampling, with the same world model and rollout budget. 
These results highlight the value of uncertainty-guided rollout allocation for policy optimization with world models.

\begin{figure*}[t]
    \centering
    \includegraphics[width=\textwidth]{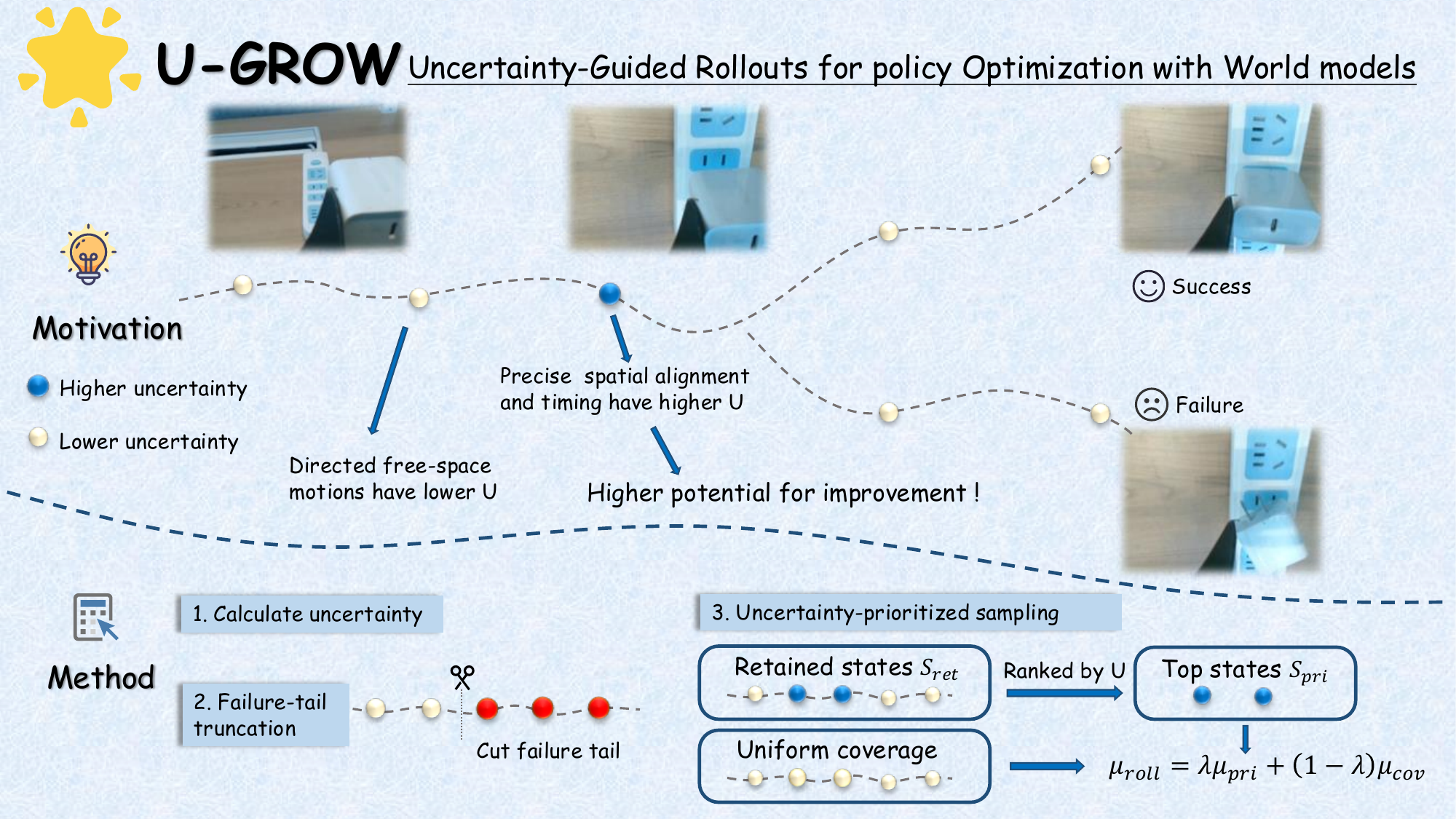}
    \caption{Overview of U-GROW}
    \label{fig:main}
\end{figure*}

\section{Related Works}
\subsection{Reinforcement Learning for VLA}
Reinforcement learning (RL) can improve the performance and generalization of Vision-Language-Action (VLA) policies beyond offline imitation~\citep{guo2025improving}.
For autoregressive VLA policies such as OpenVLA~\citep{kim2024openvla}, recent work develops on-policy RL frameworks for post-training~\citep{lu2025vla,li2026simplevla} and examines their effects on semantic generalization and execution robustness~\citep{liu2026can}.
For flow-matching policies~\citep{black2025pi05,black2024pi0}, $\pi_\texttt{RL}$ and FPO enable RL fine-tuning through stochastic denoising formulations and flow-matching-based surrogate objectives, respectively~\citep{chen2025pirl,mcallister2026flow}.
These online approaches rely on repeated environment interaction, which is costly and difficult to parallelize when training on physical robots~\citep{guo2026vlaw}.
Offline model-free RL reduces this interaction burden by learning directly from existing datasets~\citep{levine2020offline}.
Recent methods leverage value estimates to learn from data of varying quality; for example, HVD uses hierarchical value decomposition for whole-body control~\citep{zhang2026hierarchical}.
Beyond purely offline training, \citet{intelligence2025pi} use advantage conditioning to combine offline pretraining with iterative learning from demonstrations, autonomous experience, and expert corrections.
Without an explicit dynamics model to generate additional experience, however, policy improvement in the offline setting remains strongly dependent on the coverage of the available data~\citep{levine2020offline}.

\subsection{World Models for Reinforcement Learning}
Model-based reinforcement learning (MBRL) improves data efficiency by using learned dynamics to generate experience for policy optimization~\citep{janner2019mbpo,yu2020mopo,sun2023model,lin2025any,lin2026adm}.
Recent work extends this paradigm to VLA post-training through action-conditioned visual rollouts and task feedback~\citep{hou2026world,xiao2025worldenv,li2025vlarft,zhu2026wmpo}.
WMPO optimizes VLA policies through closed-loop interaction with a learned world model, using the resulting full-horizon trajectories for policy updates~\citep{zhu2026wmpo}.
A central challenge in this setting is that model prediction errors can compound over long rollouts, compromising the reliability of generated experience~\citep{janner2019mbpo}. 
To mitigate this issue, WoVR initializes rollouts from keyframes~\citep{jiang2026wovr}, while VLA-MBPO uses short-horizon, chunk-level rollouts branched from offline states~\citep{zhang2026vlambpo}.
However, model-based VLA post-training remains computationally demanding~\citep{zhu2026wmpo,zhang2026vlambpo}, motivating more effective strategies for allocating computational resources.

\section{Preliminaries}
\subsection{Markov Decision Process}
We consider a Markov decision process (MDP)
$\mathcal{M}=(\mathcal{S},\mathcal{A},T,\rho_0,\gamma)$,
where $\mathcal{S}$ and $\mathcal{A}$ denote the state and action
spaces, respectively,
$T(s_{t+1},r_t\mid s_t,a_t)$ denotes the joint distribution of the
next state and reward,
$\rho_0$ is the initial-state distribution,
and $\gamma\in[0,1)$ is the discount factor.
The objective is to learn a policy $\pi(a\mid s)$ that maximizes
the expected discounted return:
\begin{equation*}
    J(\pi)
    =
    \mathbb{E}_{\rho_0,\pi,T}
    \left[
        \sum_{t=0}^{\infty}\gamma^t r_t
    \right].
\end{equation*}

\subsection{Offline Model-based Reinforcement Learning}
In offline MBRL, the agent only has access to a fixed dataset $\Doff=\{(s_t,a_t,r_{t},s_{t+1})\}$ collected by a behavior policy. A dynamics model $\hat{T}_\phi(s_{t+1}\mid s_t,a_t)$ and a reward model $\hat{r}_\phi(s_t,a_t)$ are learned from $\Doff$ via supervised learning. MBRL methods often perform $k$-step branched rollouts with $\hat{T}_\phi$ starting from states $s\in\Doff$, store the generated transitions in a buffer $\Dwm$, and update the policy $\pi(a\mid s)$ using samples from $\Dwm$ \citep{janner2019mbpo,yu2020mopo}.

\subsection{Self-Consistency Uncertainty Estimation}
\label{sec:self-consistency-uncertainty}
Following \citet{ren2026when}, we estimate state-wise uncertainty from sensitivity to denoising discretization. For an ideal straight flow, the same state $s$ and initial noise $z$ should yield an invariant action chunk across denoising step counts; disagreement therefore indicates local policy instability.

Let $A_\theta^{(k)}(s,z)\in\mathbb{R}^{H\times D}$ denote the
action chunk generated with $k$ denoising steps, where $H$ is
the action horizon and $D$ is the action dimension.
For a set of $N\geq 2$ denoising step counts
$\mathcal{K}=\{k_1,\ldots,k_N\}$, let
$\bar A_{h,j}(s,z)=\frac{1}{N}\sum_{n=1}^{N}
A_{\theta,h,j}^{(k_n)}(s,z)$
denote the element-wise mean.
For $p\geq 1$, we define the normalized element-wise disagreement as

\begin{equation}
    u^{(p)}_{h,j}(s,z)
    =
    \frac{
      \left[
      \frac{1}{N}
      \sum_{n=1}^{N}
      \left|
          A_{\theta,h,j}^{(k_n)}(s,z)-\bar A_{h,j}
      \right|^p
      \right]^{1/p}
    }{
      \left[
      \frac{1}{N}
      \sum_{n=1}^{N}
      \left|
          A_{\theta,h,j}^{(k_n)}(s,z)
      \right|^p
      \right]^{1/p}
    },
    \label{eq:element_uncertainty}
\end{equation}
Averaging over action timesteps and dimensions yields a scalar
uncertainty score:
\begin{equation}
    U^{(p)}(s,z)
    =
    \frac{1}{HD}
    \sum_{h=1}^{H}
    \sum_{j=1}^{D}
    u^{(p)}_{h,j}(s,z).
    \label{eq:chunk_uncertainty}
\end{equation}

\section{Analyzing Policy Uncertainty in Embodied Manipulation}
\label{sec:uncertainty-analysis}

We first examine whether policy uncertainty can identify states with greater potential for policy improvement. To probe this, we conduct a policy-switching intervention on the Blocks Ranking RGB task in RoboTwin 2.0.
We obtain the base policy $\pi_{\mathrm{base}}$ by supervised fine-tuning $\pi_{0.5}$ on expert demonstrations, and further fine-tune it with RL to obtain the stronger policy $\pi^\star$. 
During evaluation, $\pi_{\mathrm{base}}$ controls the robot by default. To decide whether to switch to $\pi^\star$ at each encountered state, we estimate policy uncertainty under $\pi_{\mathrm{base}}$ using the self-consistency estimator $U(s_t,z_t)$ defined in Equation~\ref{eq:chunk_uncertainty}. 
We then compare two intervention conditions. In the high-$U$ condition, $\pi^\star$ replaces $\pi_{\mathrm{base}}$ at states satisfying $U(s_t,z_t)>\kappa_H$; in the low-$U$ condition, it replaces $\pi_{\mathrm{base}}$ at states satisfying $U(s_t,z_t)<\kappa_L$. We refer to these conditions as high-$U$ and low-$U$ switching, respectively. 
We vary the thresholds to obtain a range of replacement rates, defined as the fraction of encountered states at which $\pi^\star$ is used, and evaluate each setting over 100 episodes.

\begin{wrapfigure}{r}{0.56\textwidth}
    % \vspace{-6pt}
    \centering
    \captionsetup{skip=3pt}
    \captionsetup[subfigure]{font=scriptsize,skip=0pt}
    \begin{subfigure}[t]{0.495\linewidth}
        \centering
        \includegraphics[width=\linewidth]{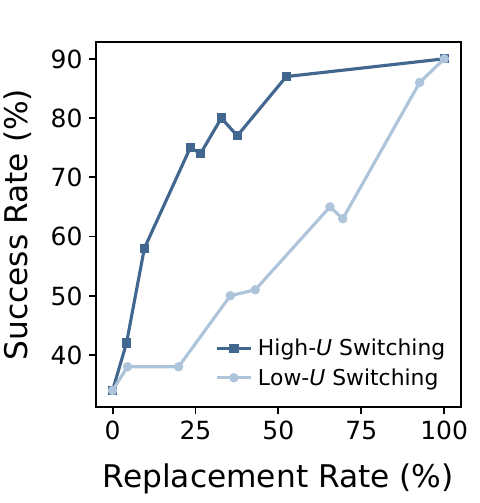}
        \caption{}
        \label{fig:policy-switching}
    \end{subfigure}\hfill
    \begin{subfigure}[t]{0.495\linewidth}
        \centering
        \includegraphics[width=\linewidth]{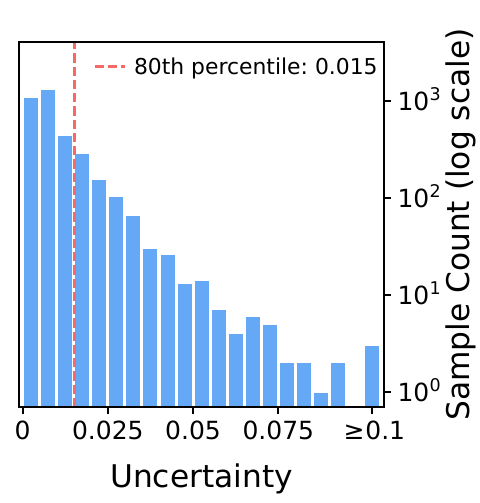}
        \caption{}
        \label{fig:uncertainty-distribution}
    \end{subfigure}
    \caption{
    % Uncertainty on Blocks Ranking RGB.
    \textbf{(a)} Success rate when a stronger policy replaces the behavior policy at high- or low-uncertainty states.
    \textbf{(b)} Uncertainty across 3,532 visited states; the dashed line marks the 80th percentile.}
    \label{fig:uncertainty-blocks-ranking}
    % \vspace{-6pt}
\end{wrapfigure}

\begin{figure*}[t]
    \centering
    \includegraphics[width=\textwidth]{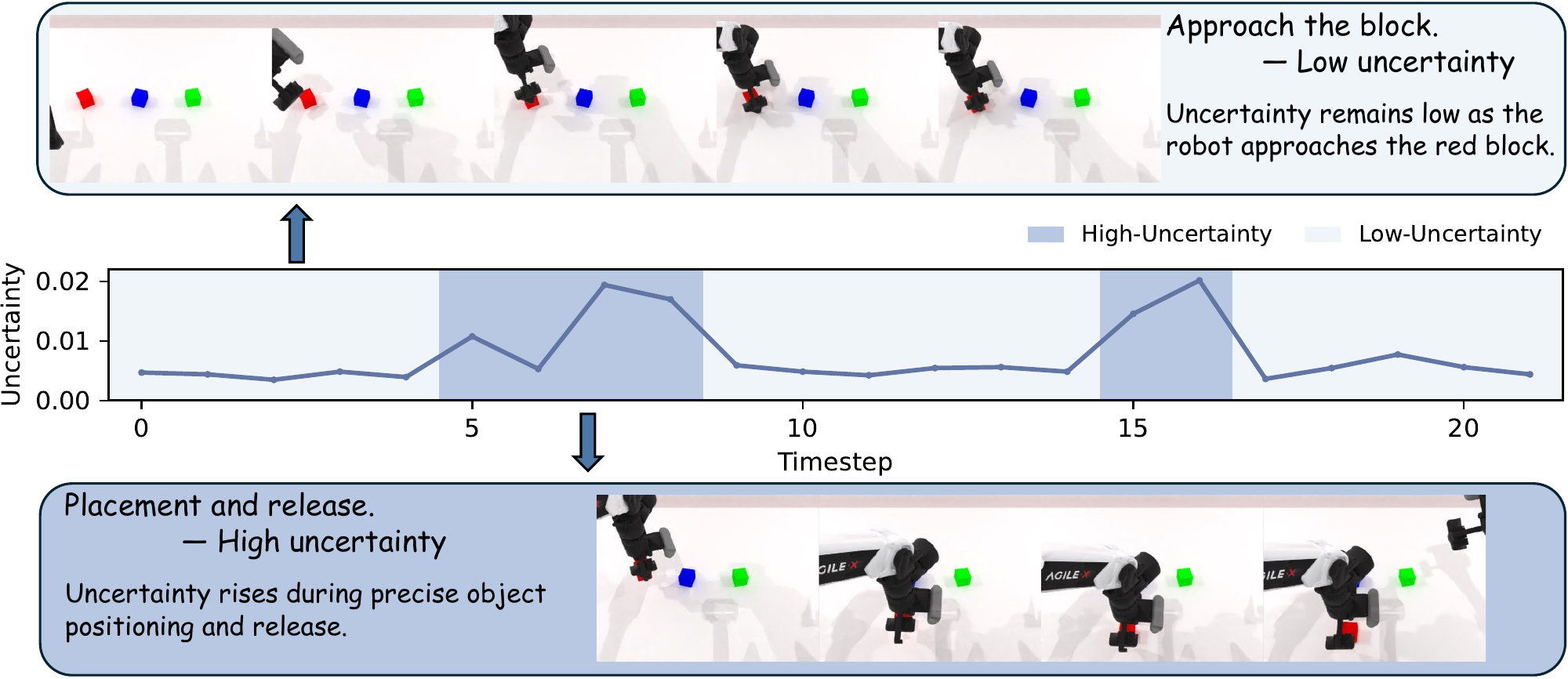}
    \caption{\textbf{Policy uncertainty across manipulation phases.}
    The curve shows policy uncertainty $U(s_t,z_t)$ along a representative trajectory. Shaded regions mark low- and high-uncertainty segments, with corresponding snapshots shown at the top and bottom, respectively. Uncertainty remains low as the robot approaches the red block but rises during placement and release, which require precise spatial alignment and timing.}
    \label{fig:uncertainty-semantics}
\end{figure*}

Figure~\ref{fig:policy-switching} shows that high-$U$ switching achieves higher success rates than low-$U$ switching across a broad range of replacement rates. At comparable replacement rates of roughly one-third, high-$U$ switching achieves an $80\%$ success rate, compared with $50\%$ for low-$U$ switching.
The larger gains from stronger-policy interventions at high-uncertainty states suggest that policy uncertainty can help identify states with greater potential for policy improvement.
This motivates using uncertainty to guide rollout-start selection.

To better understand the gains from high-$U$ switching, we characterize the distribution of uncertainty across states visited by $\pi_{\mathrm{base}}$. We rollout $\pi_{\mathrm{base}}$ for 100 episodes, obtaining 3,532 visited states, and compute $U(s_t,z_t)$ at each state. 
As shown in Figure~\ref{fig:uncertainty-distribution}, $\pi_{\mathrm{base}}$ exhibits low uncertainty at most visited states and high uncertainty at only a small fraction of them.
The 80th percentile is $0.0152$, while the maximum reaches $0.1668$.
Together with the policy-switching results, this pattern suggests that opportunities for effective policy correction are also concentrated, helping explain how high-$U$ switching achieves substantial gains at a modest replacement rate.

We next examine how policy uncertainty varies across manipulation phases.
We observe that uncertainty tends to be low during directed free-space motion but rises during grasping, placement, and release.
Figure~\ref{fig:uncertainty-semantics} illustrates this pattern in a representative trajectory: uncertainty remains low as the robot approaches the red block but increases during placement and release, which require mapping fine-grained geometric and temporal information to precise actions.
Intuitively, higher uncertainty may indicate that the current policy has not yet learned this mapping reliably.
Yet reliable action selection is particularly important here, as small action differences can alter subsequent object interactions and task outcomes. 
This offers a qualitative explanation for why high-$U$ switching is effective: it directs stronger-policy interventions toward decision-sensitive states where the base policy may struggle and better actions can have a greater impact on task success.

\begin{figure*}[t]
    \centering
    \includegraphics[width=\textwidth]{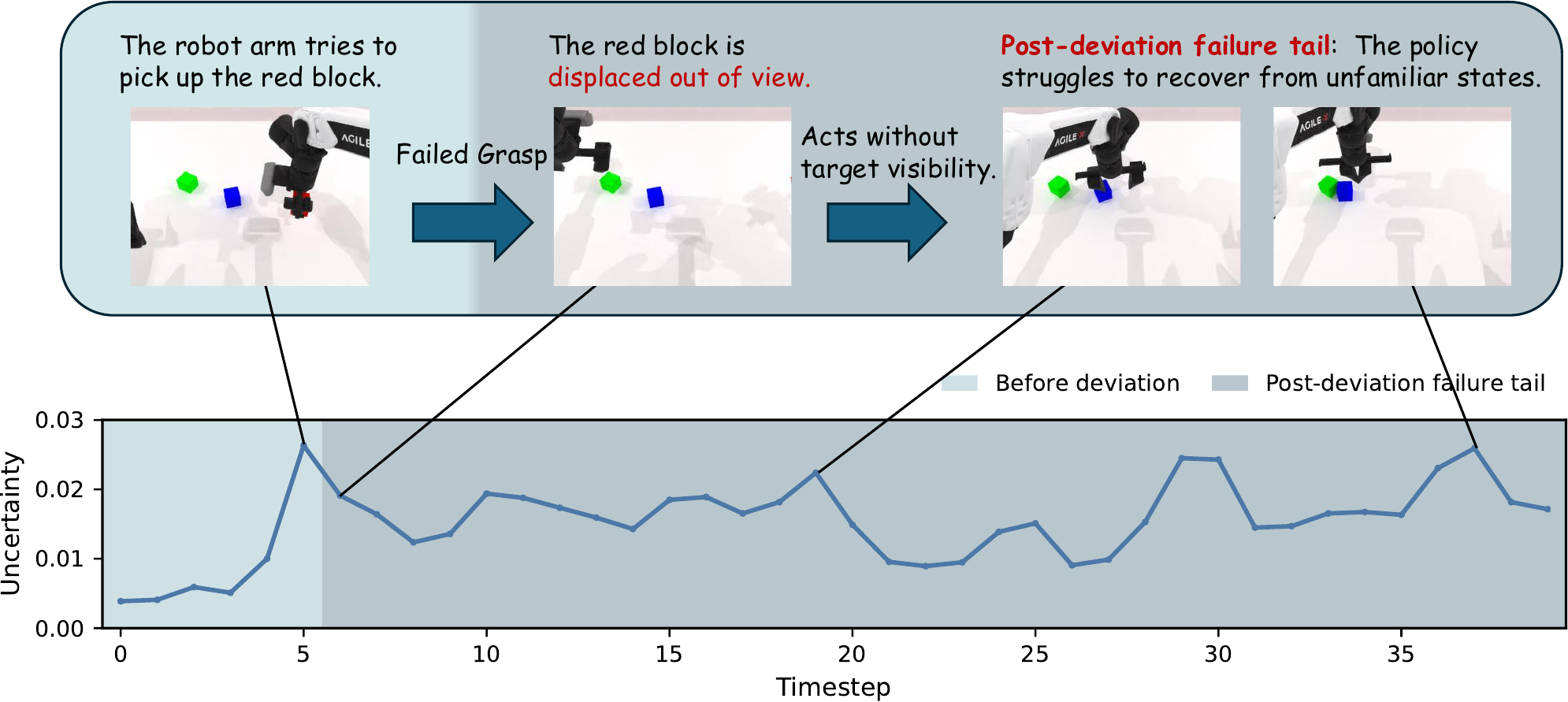}
    \caption{\textbf{Policy uncertainty in a post-deviation failure tail.}
The curve shows policy uncertainty $U(s_t,z_t)$ along a representative failed trajectory.
Snapshots illustrate a failed grasp that displaces the red block outside the camera's field of view, making recovery difficult.
The shaded region marks the resulting post-deviation failure tail, in which the policy exhibits high uncertainty at several states.
}
    \label{fig:uncertainty-failure-tail}
\end{figure*}

However, high policy uncertainty does not always indicate greater potential for policy improvement.
In failed trajectories, an early error can lead the policy into unfamiliar states from which recovery is difficult.
We refer to the subsequent trajectory segment as a \emph{post-deviation failure tail}.
Figure~\ref{fig:uncertainty-failure-tail} shows an example in which a grasping error displaces the red block outside the camera's field of view, making recovery difficult.
Although the policy exhibits high uncertainty at several states in the resulting tail, these states may be less useful as rollout starts for policy improvement.
These observations motivate excluding such failure tails from the prioritized rollout-start pool.

\section{Method}

Building on the observations above, we propose \textbf{\method{}} (\textbf{U}ncertainty-\textbf{G}uided \textbf{R}ollouts for policy \textbf{O}ptimization with \textbf{W}orld models), a sampling layer that modifies only the rollout-start distribution without changing the policy optimization objective.
Following VLA-MBPO~\citep{zhang2026vlambpo}, we generate short-horizon branched rollouts and use the resulting experience for policy updates with proximal policy optimization (PPO)~\citep{schulman2017proximal}.
To select rollout starts, \method{} first filters out failure tails to construct a candidate pool for prioritized sampling, then prioritizes high-uncertainty states within this pool.
We combine this prioritized sampling with uniform sampling over all offline states to broaden the state coverage for PPO critic training.

Let $\Doff=\{\tau_n\}_{n=1}^{N}$ denote a set of outcome-labeled offline trajectories. Each trajectory $\tau\in\Doff$ contains a state sequence $\{s_{\tau,t}\}_{t=0}^{T_\tau}$ and has a binary outcome label $y_\tau\in\{0,1\}$, where $y_\tau=1$ indicates success and $y_\tau=0$ indicates failure. For each state, we draw $z_{\tau,t}\sim p_0$ and compute \[ u_{\tau,t} = U(s_{\tau,t},z_{\tau,t}), \] where $U$ is the self-consistency uncertainty estimator defined in Equation~\ref{eq:chunk_uncertainty}. Using trajectory outcomes and uncertainty scores, \method{} first truncates post-deviation failure tails, and then prioritizes high-uncertainty states among the remaining rollout-start candidates.

\paragraph{Failure-tail truncation.} 
Failed trajectories may contain long tails after an early action error makes recovery unlikely. Based on the observed association between high uncertainty and reduced action reliability, we conservatively set the cutoff at the first high-uncertainty state and truncate the subsequent tail. We define high uncertainty using the following dataset-level quantile threshold:
\begin{equation*} 
    \kappa_{\mathrm{fail}} 
    = 
    Q_{1-\rho_{\mathrm{fail}}} \left( \left\{ u_{\tau,t} \;\middle|\; \tau\in\Doff,\; 0\leq t\leq T_\tau \right\} \right), 
    \label{eq:failure-threshold} 
\end{equation*} 
where $Q_q$ denotes the empirical $q$-quantile, and $\rho_{\mathrm{fail}} \in(0,1) $ controls the fraction of states regarded as high-uncertainty.
For each failed trajectory $\tau\in\Doff$ with $y_\tau=0$, we define its cutoff as the first timestep whose uncertainty reaches the threshold:
\begin{equation*}
    t_{\mathrm{cut}}(\tau)
    =
    \begin{cases}
        \displaystyle
        \min
        \left\{
        t
        \;\middle|\;
        0\leq t\leq T_\tau,\;
        u_{\tau,t}\geq\kappa_{\mathrm{fail}}
        \right\},
        & \text{if the set is nonempty},\\[2mm]
        T_\tau,
        & \text{otherwise}.
    \end{cases}
    \label{eq:failure-cutoff}
\end{equation*}

We then define the retained rollout-start pool as
\begin{equation*} 
\begin{aligned} 
\mathcal{S}_{\mathrm{ret}} ={}& \left\{ s_{\tau,t} \;\middle|\; \tau\in\Doff,\; y_\tau=1,\; 0\leq t\leq T_\tau \right\} \\ &\cup \left\{ s_{\tau,t} \;\middle|\; \tau\in\Doff,\; y_\tau=0,\; 0\leq t\leq t_{\mathrm{cut}}(\tau) \right\}. 
\end{aligned} 
\label{eq:retained-state-pool} 
\end{equation*} 
Thus, all states from successful trajectories remain eligible, whereas each failed trajectory contributes only the prefix preceding its post-deviation tail. 

\paragraph{Uncertainty-prioritized sampling.} 
After truncating failure tails, we rank the states in $\mathcal{S}_{\mathrm{ret}}$ by uncertainty. Let $\mathcal{S}_{\mathrm{pri}}\subseteq\mathcal{S}_{\mathrm{ret}}$ denote the top $\rho_{\mathrm{pri}}\in(0,1]$ fraction of retained states. We define the
prioritized rollout-start distribution as
\begin{equation*}
    \mu_{\mathrm{pri}}
    =
    \operatorname{Uniform}
    \left(
    \mathcal{S}_{\mathrm{pri}}
    \right).
    \label{eq:cusp-prioritized}
\end{equation*}

To broaden the state coverage of model-generated experience for PPO critic training, we retain a uniform component over all offline states:
\begin{equation*}
    \mu_{\mathrm{cov}}
    =
    \operatorname{Uniform}
    \left(
    \left\{
    s_{\tau,t}
    \;\middle|\;
    \tau\in\Doff,\;
    0\leq t\leq T_\tau
    \right\}
    \right).
    \label{eq:cusp-coverage}
\end{equation*}
The final rollout-start distribution is
\begin{equation*}
    \mu_{\mathrm{roll}}
    =
    \lambda\mu_{\mathrm{pri}}
    +
    (1-\lambda)\mu_{\mathrm{cov}},
    \qquad
    0<\lambda<1,
    \label{eq:cusp-mixture}
\end{equation*}

where $\lambda$ is the mixture weight for prioritized sampling.
We initialize short-horizon branched rollouts at states sampled from $\mu_{\mathrm{roll}}$ and use the resulting experience to update the policy and critic with PPO.

\section{Experiments}
We conduct extensive experiments on both simulated and real-world robotic manipulation tasks to answer the following questions:
(1) Does \method{} improve both sample efficiency and final policy performance in simulation?
(2) Do these gains also hold in real-world manipulation?
(3) How well do the resulting policies generalize to distribution shifts?
(4) How does each component of \method{} contribute to its overall performance?

\subsection{Simulation Task Experiments}

\paragraph{Benchmarks.} 
We evaluate \method{} on RoboTwin 2.0~\citep{chen2025robotwin} and LIBERO~\citep{liu2023libero}.
RoboTwin 2.0 comprises 50 bimanual manipulation tasks across multiple robot embodiments and supports structured domain randomization over clutter, lighting, backgrounds, tabletop height, and language instructions.
Our experiments focus on four of these tasks: \textit{Beat Block Hammer} (BH), \textit{Blocks Ranking RGB} (BR), \textit{Handover Block} (HB), and \textit{Move Can Pot} (MC).
The LIBERO evaluation spans all four suites, covering language-conditioned manipulation with variations in spatial layouts, object types, task goals, and their combinations.
We report success rates over 100 evaluation rollouts per RoboTwin 2.0 task and 500 evaluation rollouts per LIBERO suite. 

\begin{figure*}[t]
    \centering
    \includegraphics[width=\textwidth]{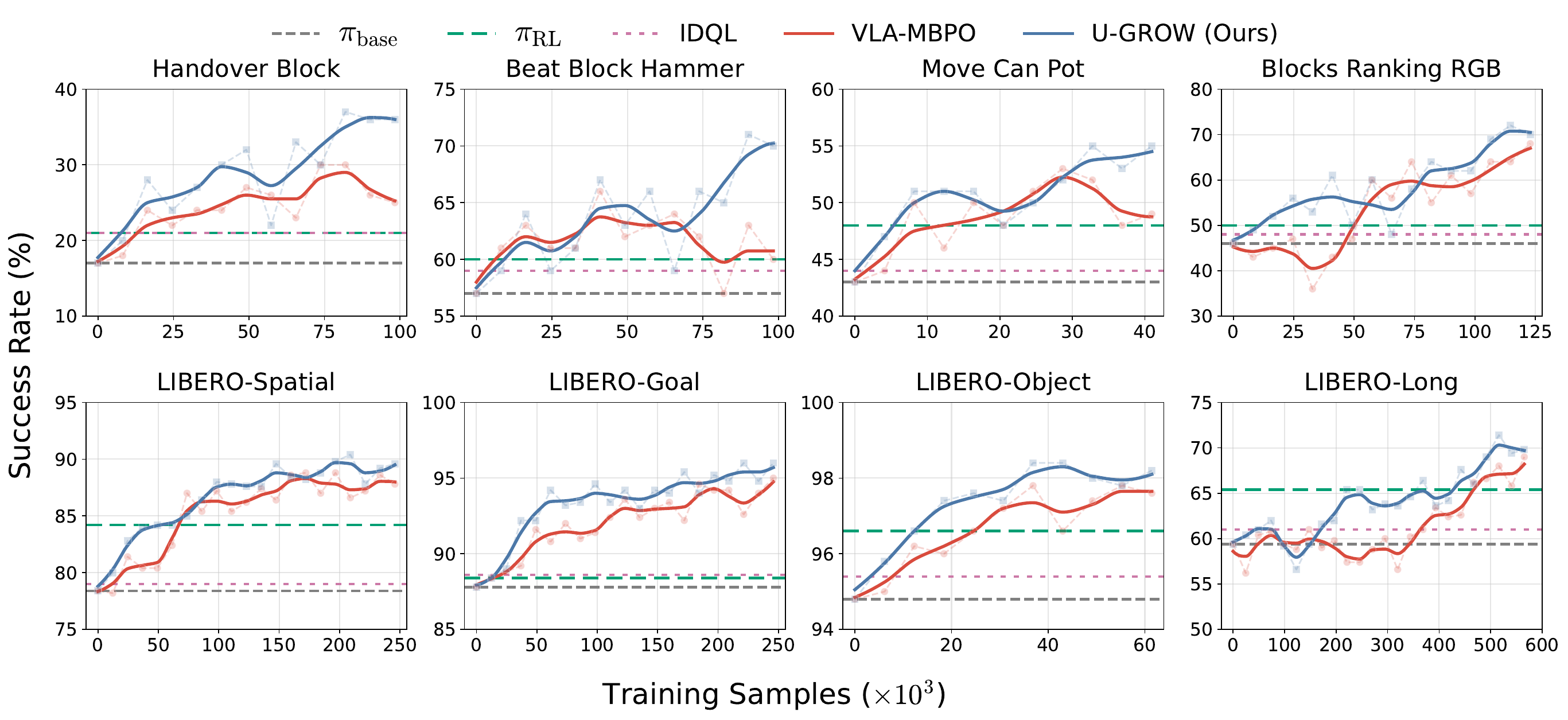}
    \caption{
    Success rate versus the cumulative number of training samples on RoboTwin 2.0 and LIBERO.
Each training sample is a transition generated by the world model.
Both methods use the same world model and total rollout budget; only the rollout-start distribution differs.}
    \label{fig:main_experiment}
\end{figure*}

\begin{table}[t]
  \centering
  \caption{Main Results.}
  \label{tab:simulation_performance_original}
  \setlength{\tabcolsep}{2pt}
    \begin{tabular}{l*{10}{C{0.07\linewidth}}}
    \toprule
    & \multicolumn{5}{c}{RoboTwin 2.0}
    & \multicolumn{5}{c}{LIBERO} \\
    \cmidrule(lr){2-6} \cmidrule(lr){7-11}
    Method &
    $\text{BH}$ & $\text{BR}$ & $\text{HB}$ & $\text{MC}$ & $\text{Avg.}$ & 
    $\text{Spatial}$ & $\text{Object}$ & $\text{Goal}$ & $\text{Long}$ & $\text{Avg.}$ \\
    \midrule
    $\pi_\mathrm{base}$
    & 57.0 & 46.0 & 17.0 & 43.0 & 40.8
    & 78.4 & 94.8 & 87.8 & 59.4 & 80.1 \\
    $\pi_\mathrm{RL}$ 
    & 60.0 & 50.0 & 21.0 & 48.0 & 44.8
    & 84.2 & 96.6 & 88.4 & 65.4 & 83.7 \\
    IDQL       
    & 59.0 & 48.0 & 21.0 & 44.0 & 43.0
    & 79.0 & 95.4 & 88.6 & 61.0 & 81.0  \\
    VLA-MBPO         
    & 66.0 & 68.0 & 30.0 & 53.0 & 54.3
    & 88.8 & 97.8 & 95.0 & 69.0 & 87.7 \\
    U-GROW (Ours)           
    & \textbf{71.0} & \textbf{72.0} & \textbf{37.0} & \textbf{55.0} & \textbf{58.8}
    & \textbf{90.4} & \textbf{98.4} & \textbf{96.0} & \textbf{71.4} & \textbf{89.1} \\
    % $\Delta$    
    % & +14.0 & +26.0 & +20.0 & +12.0 & +18.0
    % & +12.0 & +3.6 & +8.2 & +12.0 & +9.0  \\
    \bottomrule
    \end{tabular}
\end{table}

\paragraph{Baselines.}
We compare \method{} with four baselines:
(1) $\pi_{\mathrm{base}}$, the base policy obtained by supervised fine-tuning $\pi_{0.5}$~\citep{black2025pi05};
(2) $\pi_{\mathrm{RL}}$, an online RL method for flow-based VLA policies, trained under the same environment-interaction budget~\citep{chen2025pirl};
(3) IDQL, an offline model-free RL baseline~\citep{hansen2023idql}; and
(4) VLA-MBPO, a model-based RL baseline that generates chunk-level branched rollouts from states sampled uniformly from the offline dataset~\citep{zhang2026vlambpo}.
For a controlled comparison with VLA-MBPO, both methods use the same initial $\pi_\mathrm{base}$ checkpoint, offline dataset, world model, RL objective, rollout horizon, and total number of model-generated transitions; the only difference is the rollout-start distribution. 
Appendix~\ref{app:implementation} details the models, training settings, and hyperparameters.

% \paragraph{Practical Implementation.}
\paragraph{\method{} improves both sample efficiency and final policy performance.}
Under the same world model and rollout budget, \method{} improves sample efficiency, generally achieving higher success rates with the same number of training samples across RoboTwin 2.0 and LIBERO (Figure~\ref{fig:main_experiment}). The advantage is particularly pronounced on long-horizon tasks such as LIBERO-Long and Handover Block. This pattern is consistent with the intuition that longer trajectories contain more intermediate states with uneven potential for policy improvement, making selective rollout allocation increasingly valuable. 
Table~\ref{tab:simulation_performance_original} further shows that \method{} achieves the highest final success rates across all evaluated RoboTwin 2.0 tasks and LIBERO suites, with average gains of 4.5 and 1.4 percentage points over VLA-MBPO, respectively.

\subsection{Real-world Task Experiments}
\label{sec:exp_real}
We next examine whether the gains of \method{} observed in simulation also hold in real-world manipulation, where noisy observations and complex contact dynamics pose additional challenges.

\begin{figure*}[t]
    \centering
    \includegraphics[width=\textwidth]{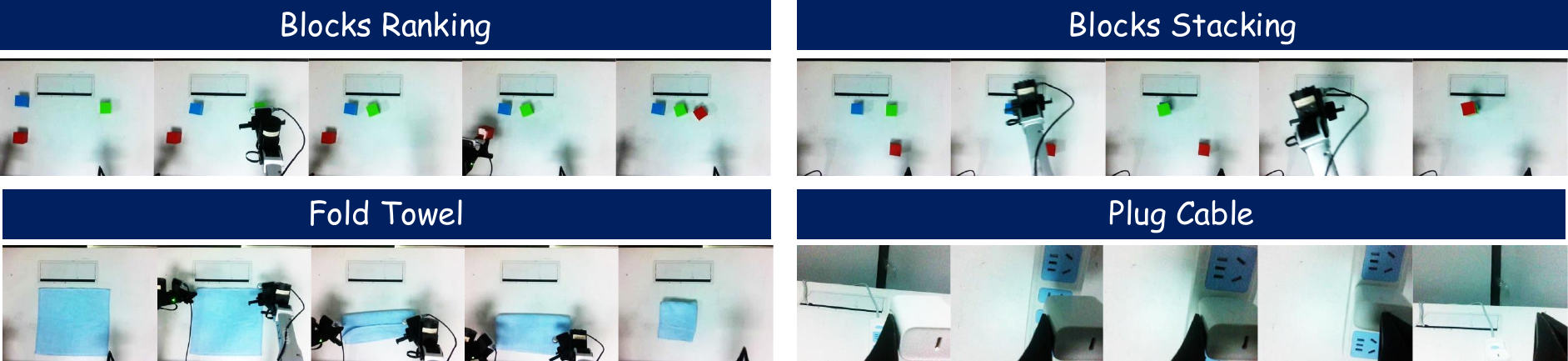}
    \caption{
Real-world manipulation tasks on the Arx-X5 bimanual platform.
\textit{Blocks Ranking} requires arranging three colored blocks according to a language instruction, while \textit{Blocks Stacking} requires stacking them in a specified order.
\textit{Fold Towel} involves bimanual folding of a deformable object, and \textit{Plug Cable} requires sub-centimeter alignment to insert a cable into a socket.
}
    \label{fig:real-world-tasks}
\end{figure*}

\paragraph{Hardware and Tasks.}
We evaluate \method{} on four real-world tasks using the Arx-X5 bimanual mobile manipulator, as illustrated in Figure~\ref{fig:real-world-tasks}. These tasks assess instruction following, contact interaction, manipulation of deformable objects, and precise insertion. 
(1) \textit{Blocks Ranking} requires arranging three colored blocks according to a language instruction; 
(2) \textit{Blocks Stacking} requires stacking the blocks in a specified order; 
(3) \textit{Fold Towel} evaluates bimanual manipulation of a deformable object; and 
(4) \textit{Plug Cable} requires sub-centimeter alignment to insert a cable into a 3-mm socket. 
For each task, we first collect 50 expert demonstrations for supervised fine-tuning of $\pi_{0.5}$, and then deploy the resulting SFT policy to collect 50 additional trajectories for world-model learning and subsequent MBRL training. 
% Each policy is evaluated over 50 rollouts per task, comprising 30 rollouts under seen configurations and 20 rollouts under unseen configurations with novel objects, backgrounds, and spatial arrangements.
Each policy is evaluated over 50 trials per task, comprising 30 under seen configurations and 20 under unseen configurations.
Appendix~\ref{app:realworld} details the hardware and control setup, data-collection procedures, and seen/unseen evaluation configurations.

\begin{figure*}[t]
    \centering
    \includegraphics[width=\textwidth]{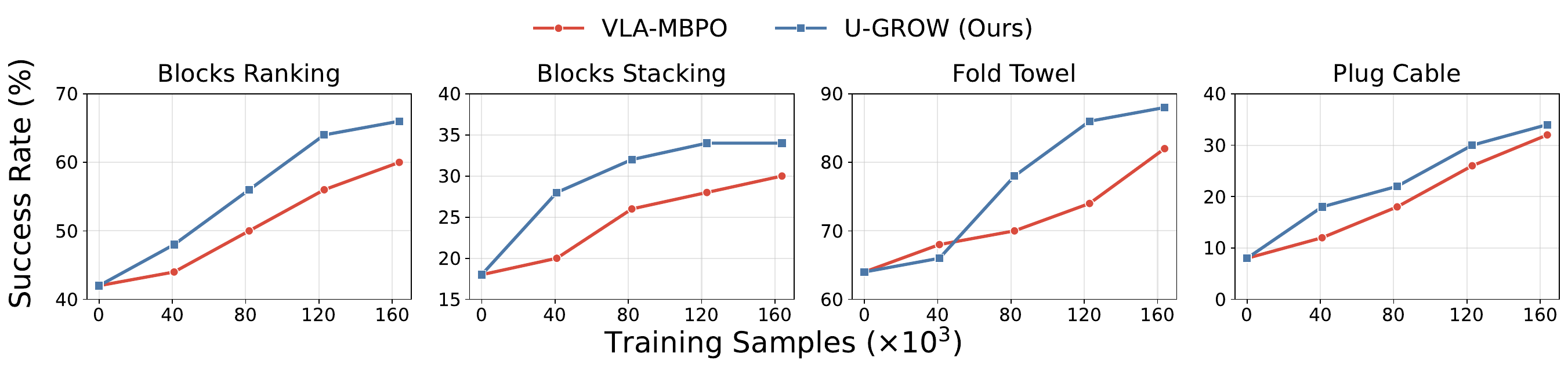}
    \caption{Success rate over the course of MBRL training on the four real-world tasks, evaluated on $50$ trials per task.}
    \label{fig:real_world_performance}
\end{figure*}

\paragraph{Results.}
As shown in Figure~\ref{fig:real_world_performance}, \method{} learns more efficiently than VLA-MBPO and achieves higher final success rates on all four real-world tasks under the same world model and rollout budget.
The gains are particularly evident on \textit{Blocks Ranking} and \textit{Fold Towel}.
The improvements span rigid-object manipulation, deformable-object handling, and precise insertion, supporting the effectiveness of uncertainty-guided rollout allocation across different types of real-world interaction.

\subsection{Generalization Evaluation}
To assess the out-of-distribution generalization of \method{}, we adopt a clean-to-randomized evaluation protocol on RoboTwin 2.0.
All methods are trained in the clean setting and evaluated in unseen randomized environments, where variations in textures, distractor objects, lighting, and tabletop geometry introduce combined visual and geometric shifts.
Additional evaluation details and representative scenes are provided in Appendix~\ref{app:simulation}.

\begin{figure*}[t]
  \centering
  \includegraphics[width=\textwidth]{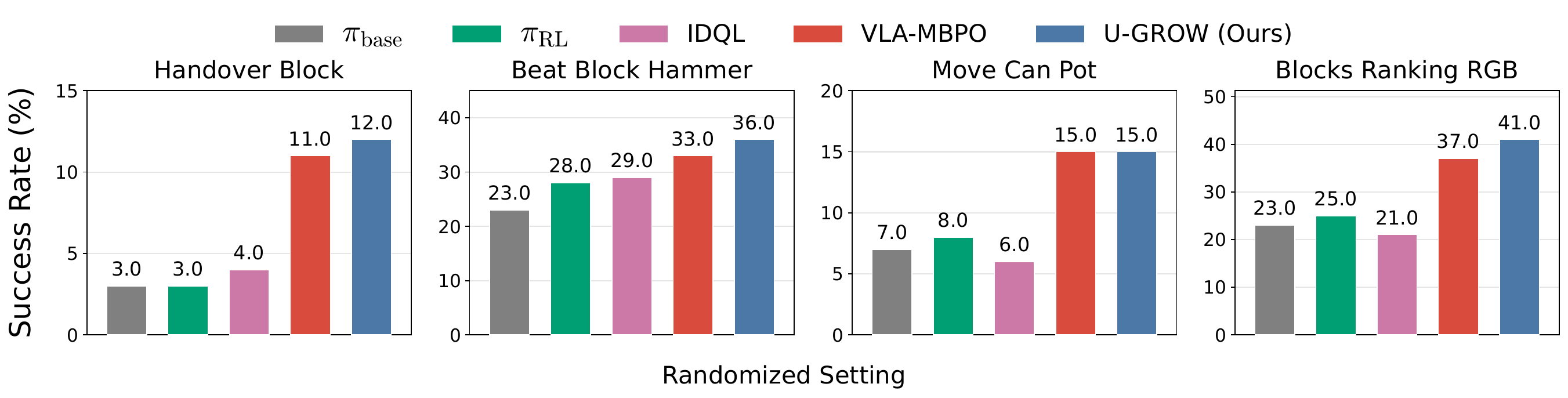}
  \caption{Success rates on the four RoboTwin 2.0 tasks in unseen randomized environments. Policies are trained in the clean setting and evaluated over $100$ rollouts per task under variations in textures, distractor objects, lighting, and tabletop geometry.}
  \label{fig:generalization_performance}
\end{figure*}

\paragraph{Results.}
As shown in Figure~\ref{fig:generalization_performance}, both MBRL methods, VLA-MBPO and \method{}, outperform the SFT policy, $\pi_{\mathrm{RL}}$, and IDQL across all four tasks in the randomized setting.
Compared with VLA-MBPO, \method{} further improves the average success rate from 24.0\% to 26.0\%, achieving higher success rates on three tasks and matching its performance on \textit{Move Can Pot}.
These results suggest that the benefits of model-based policy optimization extend beyond the clean training distribution, with additional gains from uncertainty-guided rollout allocation.

\begin{wrapfigure}{R}{0.5\columnwidth}
  \centering
  \vspace{-4pt}
  \includegraphics[width=\linewidth]{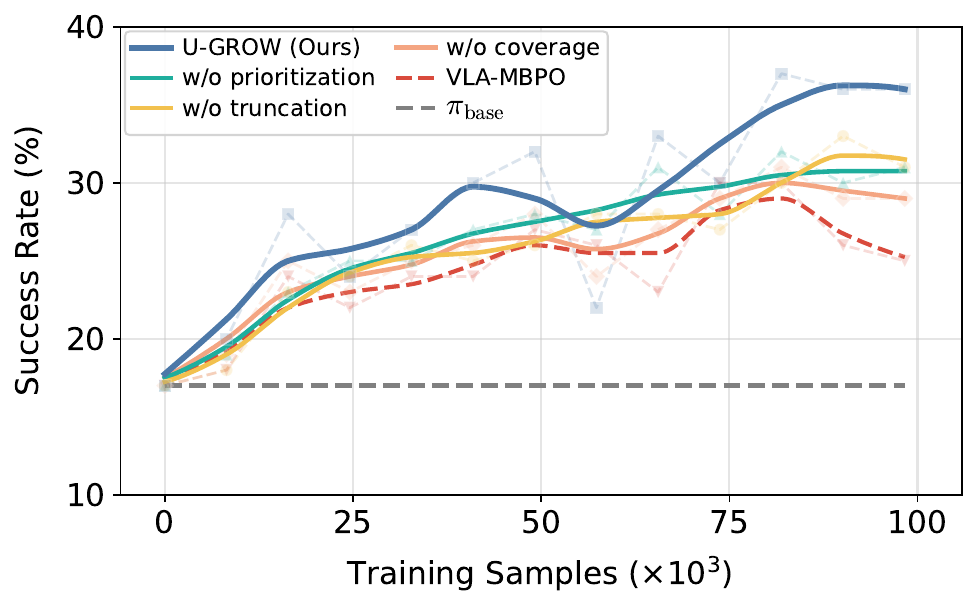}
  \caption{\method{} ablation on \textit{Handover Block}.}
  \label{fig:ablation}
  \vspace{-10pt}
\end{wrapfigure}

\subsection{Ablation Study}
To isolate the contributions of each component in \method{}, we conduct an ablation study on the \textit{Handover Block} task in RoboTwin 2.0 with three variants: 
(1) \textit{w/o truncation}, which retains all states in failed trajectories; 
(2) \textit{w/o prioritization}, which replaces the prioritized component with uniform sampling over the retained states; and 
(3) \textit{w/o coverage}, which removes the uniform component over the full offline dataset.

\paragraph{Results.}
As shown in Figure~\ref{fig:ablation}, the full method generally improves faster and achieves the highest final success rate. 
Without truncation, failure-tail states can absorb the prioritized rollout budget; without prioritization, the rollout budget is spread uniformly across retained states rather than concentrated on decision points with high-uncertainty. 
Removing coverage produces the largest performance gap, because restricting rollouts to a narrow high-utility subset reduces the state coverage available to the PPO critic, weakening value estimation over the broader policy distribution and constraining sustained policy improvement.
Together, the three components balance selective rollout allocation with the state coverage required for effective policy optimization.

\section{Conclusion}
In this work, we investigate which states offer greater potential for policy improvement and how to identify them.
Our empirical analysis suggests that policy uncertainty can help identify such states: high uncertainty is concentrated in a small subset of visited states, often during decision-sensitive manipulation stages, where stronger-policy interventions yield greater gains in task success.
Building on these findings, we introduce \method{}, a lightweight rollout allocation method for Model-based Reinforcement Learning.
After failure-tail truncation, it prioritizes high-uncertainty states as starting points for short-horizon branched rollouts, while retaining uniform sampling over all offline states to preserve broad state coverage.
Compared with uniform rollout-start sampling, \method{} improves sample efficiency and final policy performance under the same world model and rollout budget.
These results highlight the value of using policy uncertainty to direct model-generated experience toward states with greater potential for policy improvement.

% =======================================================

\bibliography{iclr2027_conference}

@inproceedings{janner2019mbpo,
  author       = {Michael Janner and
                  Justin Fu and
                  Marvin Zhang and
                  Sergey Levine},
  title        = {When to Trust Your Model: {M}odel-Based Policy Optimization},
  booktitle = {Proceedings of the 33rd Conference on Neural Information Processing Systems (NeurIPS'19), Vancouver, Canada},
  year      = {2019},
}

@inproceedings{yu2020mopo,
  author       = {Tianhe Yu and
                  Garrett Thomas and
                  Lantao Yu and
                  Stefano Ermon and
                  James Y. Zou and
                  Sergey Levine and
                  Chelsea Finn and
                  Tengyu Ma},
  title        = {{MOPO}: {M}odel-based Offline Policy Optimization},
  booktitle = {Proceedings of the 34th Conference on Neural Information Processing Systems (NeurIPS'20), Virtual Event},
  year      = {2020},
}

@inproceedings{zhang2026vlambpo,
author={Zhilong Zhang and Haoxiang Ren and Yihao Sun and Yifei Sheng and Haonan Wang and Zhichao Wu and Haoxin Lin and Pierre-Luc Bacon and Yang Yu},
title={Towards Practical World Model-based Reinforcement Learning for Vision-Language-Action Models},
booktitle = {Proceedings of the 43rd International Conference on Machine Learning (ICML'26), Seoul, South Korea},
year      = {2026},
}

@inproceedings{liu2023libero,
  author       = {Bo Liu and
                  Yifeng Zhu and
                  Chongkai Gao and
                  Yihao Feng and
                  Qiang Liu and
                  Yuke Zhu and
                  Peter Stone},
  title        = {{LIBERO}: {B}enchmarking Knowledge Transfer for Lifelong Robot Learning},
  booktitle = {Proceedings of the 37th Conference on Neural Information Processing Systems (NeurIPS'23), New Orleans, LA},
  year      = {2023},
}

@inproceedings{chen2025robotwin,
title={{RoboTwin} 2.0: {A} Scalable Data Generator and Benchmark with Strong Domain Randomization for Robust Bimanual Robotic Manipulation},
author={Tianxing Chen and Zanxin Chen and Baijun Chen and Zijian Cai and Yibin Liu and Zixuan Li and Qiwei Liang and Xianliang Lin and Yiheng Ge and Zhenyu Gu and Weiliang Deng and Yubin Guo and Tian Nian and Xuanbing Xie and Qiangyu Chen and Kailun Su and Tianling Xu and Guodong Liu and Mengkang Hu and Huan-ang Gao and Kaixuan Wang and Zhixuan Liang and Yusen Qin and Xiaokang Yang and Ping Luo and Yao Mu},
booktitle = {Proceedings of the 43rd International Conference on Machine Learning (ICML'26), Seoul, South Korea},
year      = {2026},
}

@inproceedings{ren2026when,
title={When to Trust Your Teacher: {U}ncertainty-Aware On-Policy Distillation for Embodied Policies},
author={Haoxiang Ren and Yifei Sheng and Zhilong Zhang and Runjie Xu and Bowen Lu and Haoxin Lin and Yang Yu},
booktitle={Workshop on Diffusion for Robot Learning at the 22nd Robotics: Science and Systems Conference (RSS'26), Sydney, Australia},
year={2026},
}

@inproceedings{black2024pi0,
  title = {$\pi_0$: {A} Vision-Language-Action Flow Model for General Robot Control},
  author = {Kevin Black AND Noah Brown AND Danny Driess AND Adnan Esmail AND Michael Robert Equi AND Chelsea Finn AND Niccolo Fusai AND Lachy Groom AND Karol Hausman AND Brian Ichter AND Szymon Jakubczak AND Tim Jones AND Liyiming Ke AND Sergey Levine AND Adrian Li-Bell AND Mohith Mothukuri AND Suraj Nair AND Karl Pertsch AND Lucy Xiaoyang Shi AND Laura Smith AND James Tanner AND Quan Vuong AND Anna Walling AND Haohuan Wang AND Ury Zhilinsky}, 
  booktitle = {Proceedings of the 21st Robotics: Science and Systems (RSS'25), Los Angeles, CA},
  year = {2025}
}

@inproceedings{black2025pi05,
  title = {$\pi_{0.5}$: {A} Vision-Language-Action Model with Open-World Generalization},
  author = {Black, Kevin and Brown, Noah and Darpinian, James and Dhabalia, Karan and Driess, Danny and Esmail, Adnan and Equi, Michael and Finn, Chelsea and Fusai, Niccolo and Galliker, Manuel Y. and Ghosh, Dibya and Groom, Lachy and Hausman, Karol and Ichter, Brian and Jakubczak, Szymon and Jones, Tim and Ke, Liyiming and LeBlanc, Devin and Levine, Sergey and Li-Bell, Adrian and Mothukuri, Mohith and Nair, Suraj and Pertsch, Karl and Ren, Allen Z. and Shi, Lucy Xiaoyang and Smith, Laura and Springenberg, Jost Tobias and Stachowicz, Kyle and Tanner, James and Vuong, Quan and Walke, Homer and Walling, Anna and Wang, Haohuan and Yu, Lili and Zhilinsky, Ury},
  booktitle = {Proceedings of the 9th Conference on Robot Learning (CoRL'25), Seoul, South Korea},
  year = {2025}
}

@article{chen2025pirl,
  title={$\pi_\texttt{RL}$: {O}nline {RL} fine-tuning for flow-based vision-language-action models},
  author={Kang Chen and Zhihao Liu and Tonghe Zhang and Zhen Guo and Si Xu and Hao Lin and Hongzhi Zang and Xiang Li and Quanlu Zhang and Zhaofei Yu and Guoliang Fan and Tiejun Huang and Yu Wang and Chao Yu},
  journal={arXiv preprint arXiv:2510.25889},
  year={2025}
}

@inproceedings{li2026simplevla,
  title={{SimpleVLA-RL}: {S}caling {VLA} Training via Reinforcement Learning},
  author={Haozhan Li and Yuxin Zuo and Jiale Yu and Yuhao Zhang and Zhaohui Yang and Kaiyan Zhang and Xuekai Zhu and Yuchen Zhang and Tianxing Chen and Ganqu Cui and Dehui Wang and Dingxiang Luo and Yuchen Fan and Youbang Sun and Jia Zeng and Jiangmiao Pang and Shanghang Zhang and Yu Wang and Yao Mu and Bowen Zhou and Ning Ding},
  booktitle={Proceedings of the 14th International Conference on Learning Representations (ICLR'26), Rio de Janeiro, Brazil},
  year={2026}
}

@inproceedings{liu2026can,
  title={What Can {RL} Bring to {VLA} Generalization? {A}n Empirical Study},
  author={Liu, Jijia and Gao, Feng and Wei, Bingwen and Chen, Xinlei and Liao, Qingmin and Wu, Yi and Yu, Chao and Wang, Yu},
  booktitle = {Proceedings of the 39th Conference on Neural Information Processing Systems (NeurIPS'25), San Diego, CA},
  year      = {2025},
}

@inproceedings{kim2024openvla,
  author       = {Moo Jin Kim and
                  Karl Pertsch and
                  Siddharth Karamcheti and
                  Ted Xiao and
                  Ashwin Balakrishna and
                  Suraj Nair and
                  Rafael Rafailov and
                  Ethan Paul Foster and
                  Pannag R. Sanketi and
                  Quan Vuong and
                  Thomas Kollar and
                  Benjamin Burchfiel and
                  Russ Tedrake and
                  Dorsa Sadigh and
                  Sergey Levine and
                  Percy Liang and
                  Chelsea Finn},
  title        = {{OpenVLA}: {A}n Open-Source Vision-Language-Action Model},
  booktitle    = {Proceedings of the 8th Conference on Robot Learning (CoRL'24), Munich, Germany},
  year         = {2025}
}

@inproceedings{zhang2026hierarchical,
  title={Hierarchical Value-Decomposed Offline Reinforcement Learning for Whole-Body Control},
author={Zhilong Zhang and Yunpeng Mei and Xinghao Du and Hongjie Cao and Haonan Wang and Pengyuan Min and Chenyu Wang and Pengfei Chen and Chenbo Xin and Yijie Wang and Wenyu Luo and Yihao Sun and Yidi Wang and Lei Yuan and Gang Wang and Yang Yu},
  booktitle={Proceedings of the 14th International Conference on Learning Representations (ICLR'26), Rio de Janeiro, Brazil},
  year={2026}
}

@inproceedings{intelligence2025pi,
  title={$\pi^{*}_{0.6}$: {A} {VLA} That Learns From Experience},
  author={Ali Amin AND Raichelle Aniceto AND Ashwin Balakrishna AND Kevin Black AND Ken Conley AND Grace B. Connors AND James Darpinian AND Karan Dhabalia AND Jared Di Carlo AND Danny Driess AND Michael Robert Equi AND Adnan Esmail AND Yunhao Fang AND Chelsea Finn AND Catherine Glossop AND Thomas Godden AND Ivan Goryachev AND Lachy Groom AND Hunter Hancock AND Karol Hausman AND Gashon Hussein AND Brian Ichter AND Szymon Jakubczak AND Rowan Jen AND Tim Jones AND Benjamin Katz AND Liyiming Ke AND Chandra Kuchi AND Marinda Lamb AND Devin Leblanc AND Sergey Levine AND Adrian Li-Bell AND Yao Lu AND Vishnu Mano AND Mohith Mothukuri AND Suraj Nair AND Karl Pertsch AND Allen Z. Ren AND Charvi Sharma AND Lucy Xiaoyang Shi AND Laura Smith AND Jost Tobias Springenberg AND Kyle Stachowicz AND Will Stoeckle AND Alexander Swerdlow AND James Tanner AND Marcel Torne AND Quan Vuong AND Anna Walling AND Haohuan Wang AND Blake Williams AND Sukwon Yoo AND Lili Yu AND Ury Zhilinsky AND Zhiyuan Zhou},
  booktitle = {Proceedings of the 22nd Robotics: Science and Systems (RSS'26), Sydney, Australia},
  year      = {2026},
}

@inproceedings{sun2023model,
  title={Model-Bellman Inconsistency for Model-based Offline Reinforcement Learning},
  author={Sun, Yihao and Zhang, Jiaji and Jia, Chengxing and Lin, Haoxin and Ye, Junyin and Yu, Yang},
  booktitle={Proceedings of the 40th International Conference on Machine Learning (ICML'23), Honolulu, HI},
  year={2023}
}

@inproceedings{lin2025any,
  title={Any-step Dynamics Model Improves Future Predictions for Online and Offline Reinforcement Learning},
  author={Lin, Haoxin and Xu, Yu-Yan and Sun, Yihao and Zhang, Zhilong and Li, Yi-Chen and Jia, Chengxing and Ye, Junyin and Zhang, Jiaji and Yu, Yang},
  booktitle={Proceedings of the 13th International Conference on Learning Representations (ICLR'25), Singapore},
  year={2025}
}

@inproceedings{lin2026adm,
  title={{ADM}-v2: {P}ursuing Full-Horizon Roll-out in Dynamics Models for Offline Policy Learning and Evaluation},
  author={Lin, Haoxin and Xiao, Siyuan and Li, Yi-Chen and Zhang, Zhilong and Sun, Yihao and Jia, Chengxing and Yu, Yang},
  booktitle={Proceedings of the 14th International Conference on Learning Representations (ICLR'26), Rio de Janeiro, Brazil},
  year={2026}
}

@article{xiao2025worldenv,
  title={{World-Env}: {L}everaging World Model as a Virtual Environment for {VLA} Post-Training},
  author={Xiao, Junjin and Yang, Yandan and Chang, Xinyuan and Chen, Ronghan and Xiong, Feng and Xu, Mu and Zheng, Wei-Shi and Zhang, Qing},
  journal={arXiv preprint arXiv:2509.24948},
  year={2025}
}

@article{li2025vlarft,
  title={{VLA-RFT}: {V}ision-Language-Action Reinforcement Fine-tuning with Verified Rewards in World Simulators},
  author={Hengtao Li and Pengxiang Ding and Runze Suo and Yihao Wang and Zirui Ge and Dongyuan Zang and Kexian Yu and Mingyang Sun and Hongyin Zhang and Donglin Wang and Weihua Su},
  journal={arXiv preprint arXiv:2510.00406},
  year={2025}
}

@inproceedings{zhu2026wmpo,
  title={{WMPO}: {W}orld Model-based Policy Optimization for Vision-Language-Action Models},
  author={Zhu, Fangqi and Yan, Zhengyang and Hong, Zicong and Shou, Quanxin and Ma, Xiao and Guo, Song},
  booktitle={Proceedings of the 14th International Conference on Learning Representations (ICLR'26), Rio de Janeiro, Brazil},
  year={2026}
}

@article{jiang2026wovr,
  title={{WoVR}: {W}orld models as reliable simulators for post-training {VLA} policies with {RL}},
  author={Zhennan Jiang and Shangqing Zhou and Yutong Jiang and Zefang Huang and Mingjie Wei and Yuhui Chen and Tianxing Zhou and Zhen Guo and Hao Lin and Quanlu Zhang and Yu Wang and Haoran Li and Chao Yu and Dongbin Zhao},
  journal={arXiv preprint arXiv:2602.13977},
  year={2026}
}

@article{hou2026world,
  title={World Model for Robot Learning: {A} Comprehensive Survey},
  author={Bohan Hou and Gen Li and Jindou Jia and Tuo An and Xinying Guo and Sicong Leng and Haoran Geng and Yanjie Ze and Tatsuya Harada and Philip Torr and Oier Mees and Marc Pollefeys and Zhuang Liu and Jiajun Wu and Pieter Abbeel and Jitendra Malik and Yilun Du and Jianfei Yang},
  journal={arXiv preprint arXiv:2605.00080},
  year={2026}
}

@article{hansen2023idql,
  title={{IDQL}: {I}mplicit {Q}-learning as an actor-critic method with diffusion policies},
  author={Hansen-Estruch, Philippe and Kostrikov, Ilya and Janner, Michael and Kuba, Jakub Grudzien and Levine, Sergey},
  journal={arXiv preprint arXiv:2304.10573},
  year={2023}
}

@article{agarwal2026cosmos,
  title = {Cosmos 3: {O}mnimodal World Models for Physical {AI}},
  author = {{NVIDIA}},
  journal = {arXiv preprint arXiv:2606.02800},
  year = {2026},
}

@inproceedings{guo2025improving,
  author       = {Yanjiang Guo and
                  Jianke Zhang and
                  Xiaoyu Chen and
                  Xiang Ji and
                  Yen{-}Jen Wang and
                  Yucheng Hu and
                  Jianyu Chen},
  title        = {Improving Vision-Language-Action Model with Online Reinforcement Learning},
  booktitle    = {Proceedings of the 42nd IEEE International Conference on Robotics and Automation (ICRA'25), Atlanta, GA},
  year         = {2025}
}

@inproceedings{guo2026vlaw,
  title={{VLAW}: {I}terative Co-Improvement of Vision-Language-Action Policy and World Model},
  author={Guo, Yanjiang and Lee, Tony and Shi, Lucy Xiaoyang and Chen, Jianyu and Liang, Percy and Finn, Chelsea},
  booktitle = {Proceedings of the 43rd International Conference on Machine Learning (ICML'26), Seoul, South Korea},
  year={2026}
}

@article{levine2020offline,
  title={Offline Reinforcement Learning: {T}utorial, Review, and Perspectives on Open Problems},
  author={Levine, Sergey and Kumar, Aviral and Tucker, George and Fu, Justin},
  journal={arXiv preprint arXiv:2005.01643},
  year={2020}
}

@article{zhang2024whale,
  title={{WHALE}: {T}owards Generalizable and Scalable World Models for Embodied Decision-making},
  author={Zhilong Zhang and Ruifeng Chen and Junyin Ye and Yihao Sun and Pengyuan Wang and Jingcheng Pang and Kaiyuan Li and Tianshuo Liu and Haoxin Lin and Yang Yu and Zhi-Hua Zhou},
  journal={arXiv preprint arXiv:2411.05619},
  year={2024}
}

@article{yu2026should,
  title={How Should World Models Be Evaluated for Embodied Decision-Making? {A} Decision-Making-Centric Position},
  author={Yu, Yang and Zhang, Shiyuan and Sheng, Yifei and Ren, Haoxiang and Lin, Haoxin},
  journal={arXiv preprint arXiv:2606.15032},
  year={2026}
}

@article{schulman2017proximal,
  author       = {John Schulman and
                  Filip Wolski and
                  Prafulla Dhariwal and
                  Alec Radford and
                  Oleg Klimov},
  title        = {Proximal Policy Optimization Algorithms},
  journal      = {arXiv preprint arXiv:1707.06347},
  year         = {2017}
}

@article{wan2025wanopenadvancedlargescale,
  author       = {{Team Wan} and Ang Wang and Baole Ai and Bin Wen and Chaojie Mao and Chen-Wei Xie and Di Chen and Feiwu Yu and Haiming Zhao and Jianxiao Yang and Jianyuan Zeng and Jiayu Wang and Jingfeng Zhang and Jingren Zhou and Jinkai Wang and Jixuan Chen and Kai Zhu and Kang Zhao and Keyu Yan and Lianghua Huang and Mengyang Feng and Ningyi Zhang and Pandeng Li and Pingyu Wu and Ruihang Chu and Ruili Feng and Shiwei Zhang and Siyang Sun and Tao Fang and Tianxing Wang and Tianyi Gui and Tingyu Weng and Tong Shen and Wei Lin and Wei Wang and Wei Wang and Wenmeng Zhou and Wente Wang and Wenting Shen and Wenyuan Yu and Xianzhong Shi and Xiaoming Huang and Xin Xu and Yan Kou and Yangyu Lv and Yifei Li and Yijing Liu and Yiming Wang and Yingya Zhang and Yitong Huang and Yong Li and You Wu and Yu Liu and Yulin Pan and Yun Zheng and Yuntao Hong and Yupeng Shi and Yutong Feng and Zeyinzi Jiang and Zhen Han and Zhi-Fan Wu and Ziyu Liu},
  title        = {Wan: {O}pen and Advanced Large-Scale Video Generative Models},
  journal      = {arXiv preprint arXiv:2503.20314},
  year         = {2025}
}

@misc{qwen3.5,
    title  = {{Qwen3.5}: {T}owards Native Multimodal Agents},
    author = {{Qwen Team}},
    year   = {2026},
    url    = {https://qwen.ai/blog?id=qwen3.5}
}

@article{lu2025vla,
  title={{VLA-RL}: {T}owards masterful and general robotic manipulation with scalable reinforcement learning},
  author={Lu, Guanxing and Guo, Wenkai and Zhang, Chubin and Zhou, Yuheng and Jiang, Haonan and Gao, Zifeng and Tang, Yansong and Wang, Ziwei},
  journal={arXiv preprint arXiv:2505.18719},
  year={2025}
}

@inproceedings{mcallister2026flow,
  title={Flow matching policy gradients},
  author={McAllister, David and Ge, Songwei and Yi, Brent and Kim, Chung Min and Weber, Ethan and Choi, Hongsuk and Feng, Haiwen and Kanazawa, Angjoo},
  booktitle={Proceedings of the 14th International Conference on Learning Representations (ICLR'26), Rio de Janeiro, Brazil},
  year={2026}
}
\bibliographystyle{iclr2027_conference}

\appendix
\clearpage
\providecommand{\appfill}[1]{\textit{[TBD: #1]}}
\providecommand{\appblank}{\textit{[TBD]}}

\section{Computational Resources}
All training runs are conducted on 8 NVIDIA H100 GPUs. World-model training on the collected offline dataset takes approximately 40 hours per run, while subsequent model-based policy optimization takes approximately 5-6 hours per run under the specified model-rollout budget.

\section{Experimental setup}
\label{app:protocols}

\begin{figure}[htbp]
    \centering
    \includegraphics[width=\linewidth]{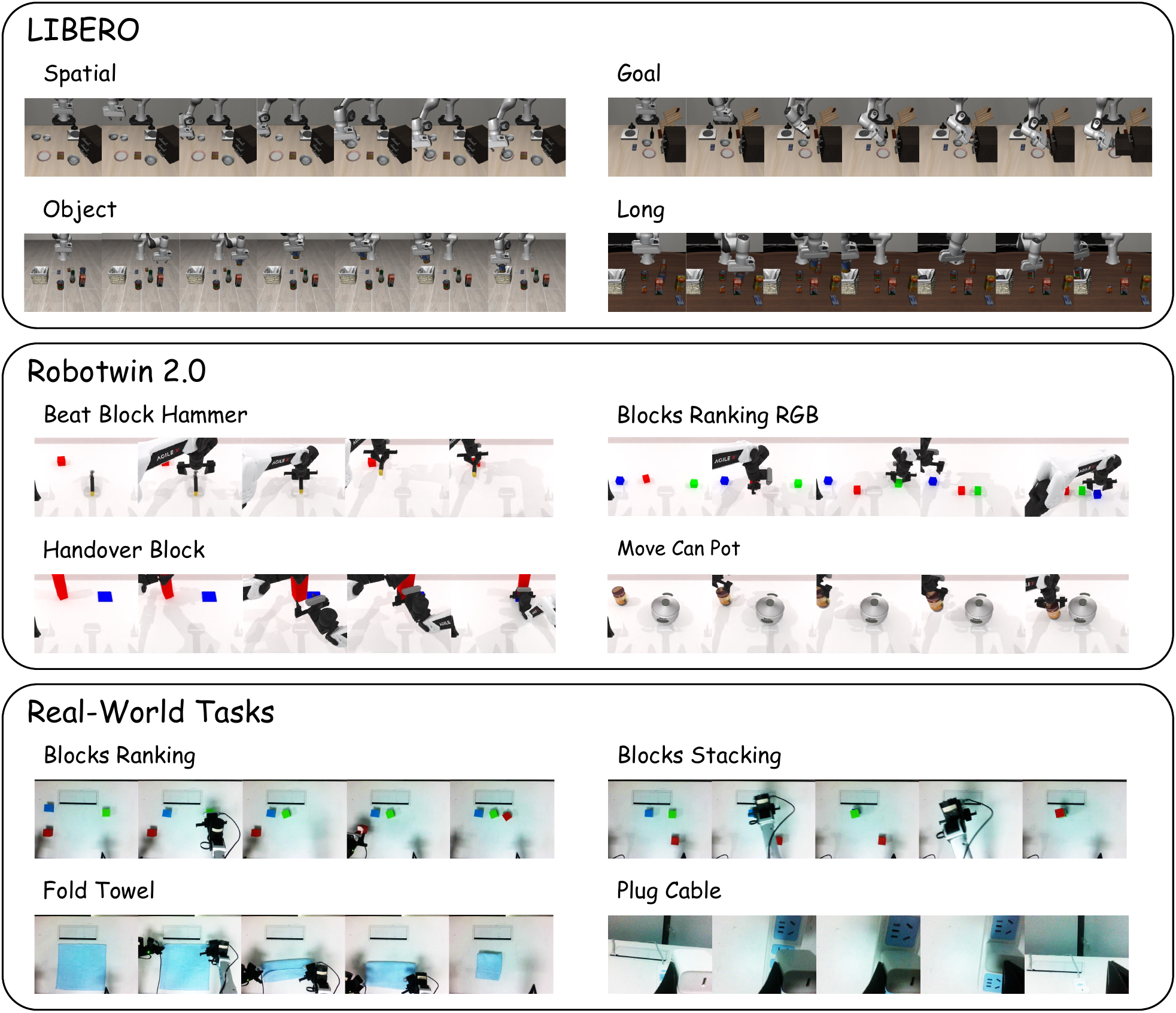}
    \caption{Representative manipulation sequences across the three experimental settings.}
    \label{fig:app-task-illustration}
\end{figure}

Figure~\ref{fig:app-task-illustration} provides an overview of the manipulation tasks used in our experiments, with representative sequences from LIBERO, RoboTwin 2.0, and the Real-World Tasks setup. 

\subsection{Simulation experiments}
\label{app:simulation}
We use four RoboTwin 2.0 tasks~\citep{chen2025robotwin}: Beat Block Hammer (BH), Blocks Ranking RGB (BR), Handover Block (HB), and Move Can Pot (MC). For LIBERO~\citep{liu2023libero}, we use four suites: Spatial, Object, Goal, and Long.
For supervised fine-tuning (SFT), we use 100 and 40 expert trajectories per task in RoboTwin 2.0 and LIBERO, respectively. We also collect an offline dataset comprising 100 trajectories per RoboTwin task and 500 trajectories per LIBERO suite. This dataset is used to train the world model and provide
starting states for branched rollouts during policy optimization. 
Each policy is evaluated over 100 episodes per RoboTwin 2.0 task
in each of the clean and randomized settings, and over 500 episodes
per LIBERO suite.

\begin{figure}[htbp]
    \centering
    \includegraphics[width=\linewidth]{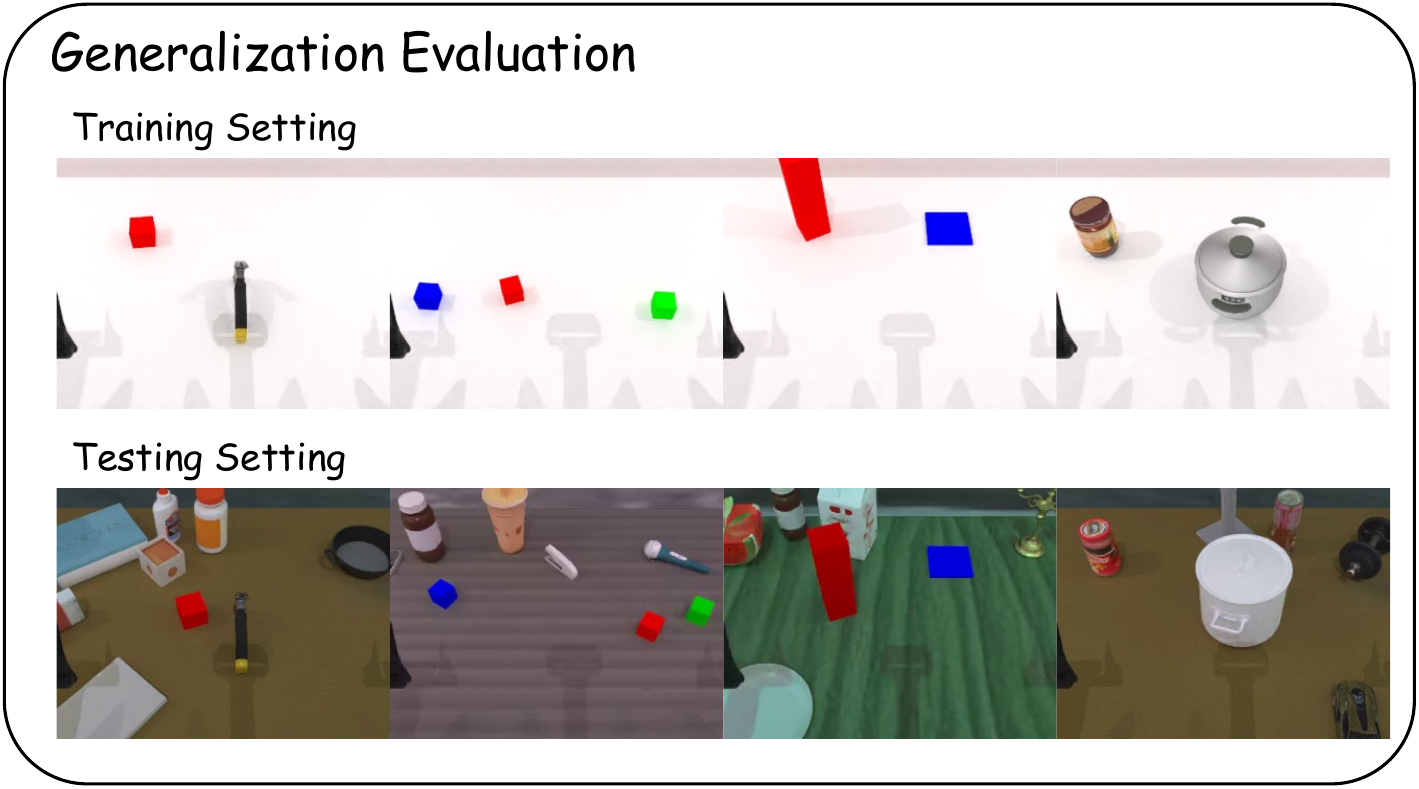}
    \caption{Clean-to-randomized evaluation on RoboTwin 2.0. The top row shows the clean training setting, and the bottom row shows representative randomized test scenes. From left to right, the columns correspond to Beat Block Hammer, Blocks Ranking RGB, Handover Block, and Move Can Pot. Policies trained in the clean setting are evaluated under visual and geometric shifts without additional training in the randomized environments.}
    \label{fig:app-generalization}
\end{figure}

\paragraph{Generalization Evaluation Setup.}
We use a clean-to-randomized protocol on RoboTwin 2.0: policies are trained in the clean setting, and the same checkpoints are evaluated in unseen randomized environments. The randomized setting varies textures, distractor objects, lighting, and tabletop geometry. Figure~\ref{fig:app-generalization} compares clean training scenes with representative randomized test scenes for all four tasks. Success rates in both settings are reported in Table~\ref{fig:generalization_performance}.

\subsection{Real-world experiments}
\label{app:realworld}

\begin{figure}[htbp]
    \centering
    \includegraphics[width=0.9\linewidth]{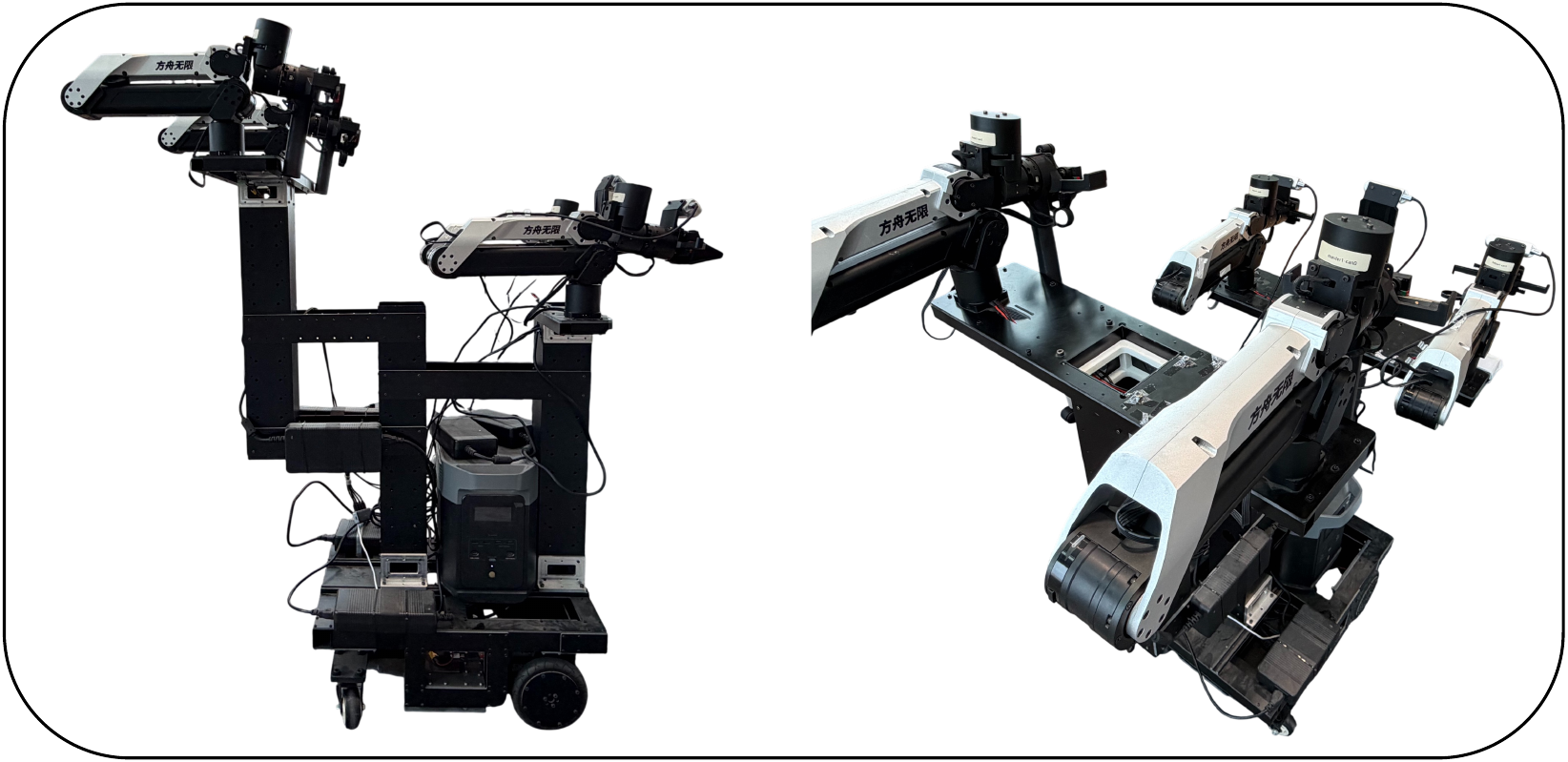}
    \caption{Arx-X5 platform for real-world data collection and policy evaluation.}
    \label{fig:app-arx-x5}
\end{figure}

\paragraph{Hardware and control.}
We use the Arx-X5 bimanual platform shown in Figure~\ref{fig:app-arx-x5} for the real-world experiments. The platform has 14 degrees of freedom and is equipped with three Intel RealSense D435i RGB-D cameras for multi-view observations. Expert demonstrations are collected using a master--slave teleoperation setup, in which two handheld master arms control the corresponding robot arms. The system operates at a control frequency of 15\,Hz.

\paragraph{Tasks.}
The lower panel of Figure~\ref{fig:app-task-illustration} shows representative execution sequences for the four real-world tasks. \textit{Blocks Ranking} requires arranging three colored blocks according to a language instruction, and \textit{Blocks Stacking} requires stacking blocks in a specified order. \textit{Fold Towel} involves folding a towel through coordinated bimanual manipulation, while \textit{Plug Cable} requires aligning a cable plug with a socket and inserting it.

\begin{figure}[htbp]
    \centering
    \includegraphics[width=\linewidth]{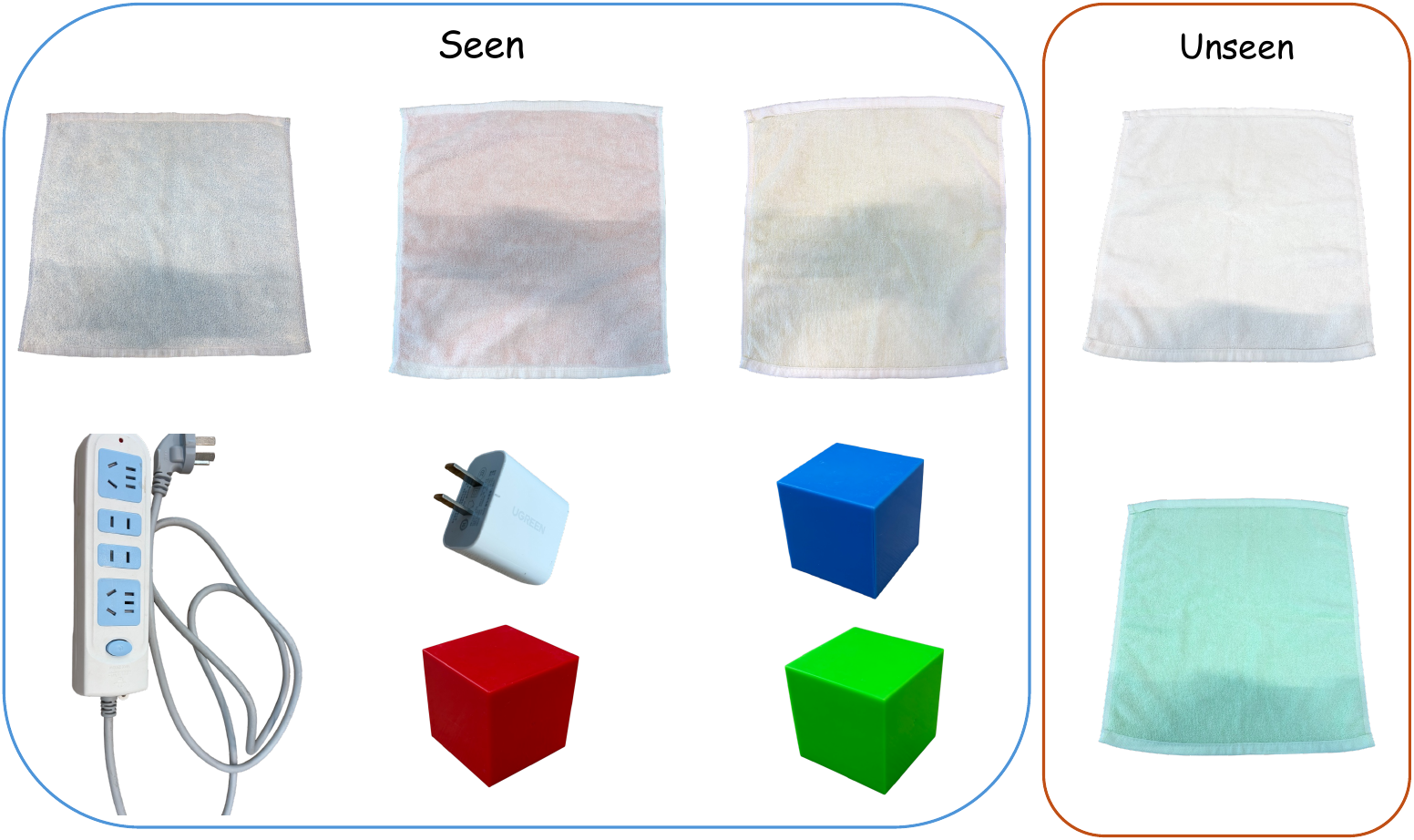}
    \caption{Objects used in the real-world experiments. 
    \textbf{Left (seen):} three towels for \textit{Fold Towel}, colored blocks for \textit{Blocks Ranking} and \textit{Blocks Stacking}, and a power strip and plug adapter for \textit{Plug Cable}. \textbf{Right (unseen):} two held-out towels for evaluating \textit{Fold Towel} on unseen object instances.}
    \label{fig:app-realworld-objects}
\end{figure}

\paragraph{Data collection and evaluation.}
For each task, we collect 50 expert demonstrations for SFT and then deploy the resulting policy to collect 50 additional trajectories. We use these trajectories to train the world model and initialize branched rollouts during policy optimization.
Each policy is evaluated over 50 trials per task, comprising 30 under seen configurations and 20 under unseen configurations. The unseen configurations introduce task-specific variations: language instructions specifying unseen block orders in \textit{Blocks Ranking} and \textit{Blocks Stacking}, unseen towel colors in \textit{Fold Towel}, and changes in object positions in \textit{Plug Cable}. Figure~\ref{fig:app-realworld-objects} shows the objects used in the seen configurations and the two held-out towels used for unseen evaluation in \textit{Fold Towel}. Figure~\ref{fig:real_world_performance} reports success rates over all 50 trials per task.

\section{Implementation details}
\label{app:implementation}

\subsection{World model}
\label{app:worldmodel}

We adapt the pretrained Wan2.2-TI2V-5B video diffusion model~\citep{wan2025wanopenadvancedlargescale} to predict $K$ future observations conditioned on five context observations, five historical actions, and a policy action chunk of length $K$.
At each time step, camera views are stacked vertically into a single image. The VAE encodes the five context images into two temporal latent slices, which are assigned diffusion timestep zero and remain fixed throughout denoising.

\paragraph{Action conditioning.}
Actions condition every DiT block through two pathways.
For \emph{action cross-attention}, a two-layer MLP embeds each action, and video tokens attend to the resulting embeddings.
For \emph{action-dependent modulation}, we prepend three zero actions to the action sequence and concatenate each consecutive group of four actions to match the VAE's temporal compression.
A separate MLP maps each group to an embedding, which is combined with the diffusion-timestep embedding to control normalization and residual gating in the DiT blocks, as well as output-head normalization.

\paragraph{Training.}
We initialize three separate world models from the base Wan weights: one trained jointly across the four LIBERO suites, one across the four RoboTwin tasks, and one across the four real-world tasks. 
For each model, we fine-tune the full DiT and both action MLPs while keeping the VAE frozen, using a timestep-weighted flow-matching MSE loss over future video latents.

The executed action chunk length is $K=20$ for RoboTwin and the real-world tasks, and $K=8$ for LIBERO. Table~\ref{tab:app-worldmodel} lists the training and rollout settings. 

\begin{table}[htbp]
\centering
\caption{World model training settings.}
\label{tab:app-worldmodel}
\small
\begin{tabular}{l c c c}
\toprule
Setting & RoboTwin 2.0 & LIBERO & Real-World Tasks\\
\midrule
Image resolution & $256\times256$ & $256\times256$ & $256\times256$ \\
Context frames & 5 & 5 & 5 \\
Video sequence length & 25 & 13 & 25 \\
Denoising steps & 5 & 5 & 5 \\
Guidance scale & 1.0 & 1.0 & 1.0 \\
Noise-schedule shift & 5.0 & 5.0 & 5.0 \\
Learning rate & 1e-5 & 1e-5 & 1e-5 \\
Batch size & 256 & 256 & 256 \\
\bottomrule
\end{tabular}
\end{table}

\subsection{Reward model}
\label{app:rewardmodel}

We fine-tune Qwen3.5-0.8B~\citep{qwen3.5} to predict task completion
with \texttt{Yes}/\texttt{No} responses.
We train a separate reward model for each LIBERO suite and each
RoboTwin or real-world task, and keep all reward models fixed
during policy optimization.

\paragraph{Inputs and supervision.}
Each example pairs a task instruction with visual observations from a single time step.
LIBERO training uses the main-camera image. RoboTwin and real-world training use three separate  images, ordered as the head, left-wrist, and right-wrist views, in a single multimodal prompt. 
Labels are derived from trajectory outcomes: for successful trajectories, the final five frames are labeled \texttt{Yes}, and earlier frames are labeled \texttt{No}. All frames from failed trajectories are negative. 
We balance the two classes by oversampling positive examples to match the negatives.

\paragraph{Training.}
We fine-tune the language model and visual alignment module while freezing the vision encoder. The objective is autoregressive cross-entropy over the assistant's answer tokens, with prompt tokens masked from the loss. Table~\ref{tab:app-rewardmodel} reports the optimization settings. 

\begin{table}[htbp]
\centering
\caption{Reward model training settings.}
\label{tab:app-rewardmodel}
\small
\begin{tabular}{l c c c}
\toprule
Setting & RoboTwin & LIBERO & Real-World Tasks \\
\midrule
Learning rate & $2\times10^{-5}$ & $2\times10^{-5}$ & $2\times10^{-5}$ \\
Weight decay & 0.1 & 0.1 & 0.1 \\
Adam $(\beta_1,\beta_2)$ & (0.9, 0.95) & (0.9, 0.95) & (0.9, 0.95) \\
Adam $\epsilon$ & $10^{-8}$ & $10^{-8}$ & $10^{-8}$ \\
Learning-rate schedule & Cosine & Cosine & Cosine \\
Warmup fraction & 0.05 & 0.05 & 0.05 \\
Gradient norm clipping & 1.0 & 1.0 & 1.0 \\
Global batch size & 256 & 256 & 256 \\
Maximum sequence length & 4096 & 4096 & 4096 \\
Precision & BF16 & BF16 & BF16 \\
\bottomrule
\end{tabular}
\end{table}

\paragraph{Rollout rewards.}
During model rollouts, the reward model evaluates each of the
$K$ generated observations independently, conditioned on the
task instruction and the camera views at that time step.
Let $z^{\mathrm{Yes}}_{c,j}$ and $z^{\mathrm{No}}_{c,j}$ denote
the logits of the \texttt{Yes} and \texttt{No} tokens at the
first answer position for time step $j$ of action chunk $c$.
We normalize these two logits to obtain the completion score
\begin{equation*}
    p_{c,j}
    =
    \frac{\exp(z^{\mathrm{Yes}}_{c,j})}
    {\exp(z^{\mathrm{Yes}}_{c,j})+\exp(z^{\mathrm{No}}_{c,j})}.
\end{equation*}
A chunk is predicted successful if any generated observation
has $p_{c,j}\geq0.5$.
We assign its binary reward to the final step, with zero rewards
at all preceding steps:
\begin{equation*}
    r_{c,j}
    =
    \mathbf{1}[j=K]\,
    \mathbf{1}\!\left[
        \max_{1\leq m\leq K}p_{c,m}\geq0.5
    \right].
\end{equation*}
An episode-level success flag remains set once any chunk is
predicted successful and is cleared on reset.
Rewards are computed independently for each chunk.

\subsection{Policy optimization}
\label{app:ppo}

We update the policy and critic with PPO using chunk-level rewards and log probabilities. The critic takes detached VLM features as input. We compute advantages using generalized advantage estimation (GAE) and normalize them within each task.
Table~\ref{tab:app-ppo} lists the optimization settings.

\begin{table}[htbp]
\centering
\caption{Policy optimization settings.}
\label{tab:app-ppo}
\small
\begin{tabular}{l c c c}
\toprule
Setting & RoboTwin & LIBERO & Real-World Tasks\\
\midrule
Action chunk & 20 & 8 & 20 \\
Action dimension & 14 & 7 & 14 \\
Denoising steps & 5 & 3 & 5 \\
Rollout chunk & 2 & 3 & 2 \\
Actor learning rate & $2.5\times10^{-6}$ & $5\times10^{-6}$ & $2.5\times10^{-6}$ \\
Critic learning rate & $5\times10^{-5}$ & $10^{-4}$ & $5\times10^{-5}$ \\
Batch size & 1024 & 1024 & 1024 \\
PPO update epochs & 5 & 10 & 5 \\
Discount Rate & 0.99 & 0.99 & 0.99 \\
GAE & 0.95 & 0.95 & 0.95 \\
Noise level & 0.5 & 0.3 & 0.5 \\
Adam $(\beta_1,\beta_2)$ & (0.9, 0.95) & (0.9, 0.95) & (0.9, 0.95) \\
\bottomrule
\end{tabular}
\end{table}

\subsection{Uncertainty-guided rollout starts}
\label{app:sampler}

We use the uncertainty estimator in Section~\ref{sec:self-consistency-uncertainty} with $p=2$ and two denoising step counts, $\mathcal{K}=\{3, 5\}$.
At each state, both denoising evaluations share the same initial noise.
Uncertainty scores are computed for all candidate offline states before rollout-start selection. We set $\rho_{\mathrm{fail}}=0.1, \rho_{\mathrm{pri}}=0.2$ and the prioritized sampling weight to $\lambda=0.5$.

\end{document}